\documentclass[10pt,twocolumn,letterpaper]{article}

\usepackage[pagenumbers]{cvpr} 

\usepackage{graphicx}
\usepackage{amsmath}
\usepackage{amssymb}
\usepackage{booktabs}
\usepackage{algorithm}
\usepackage{algpseudocode}
\usepackage{multirow}
\usepackage[dvipsnames]{xcolor,colortbl}
\usepackage{chapterbib}
\usepackage{epstopdf}
\usepackage{stackengine}
\usepackage{adjustbox}
\usepackage{setspace}

\usepackage{microtype}

\DeclareMathAlphabet{\mbf}{OT1}{ptm}{b}{n}

\newcommand{\mrm}[1]{\mathrm{#1}}
\newcommand{\trans}{{\ensuremath{\mathsf{T}}}}

\definecolor{cvprblue}{rgb}{0.21,0.49,0.74}
\usepackage[pagebackref,breaklinks,colorlinks,allcolors=cvprblue]{hyperref}
\usepackage{cuted}
\usepackage{capt-of}

\def\paperID{17030} 
\def\confName{CVPR}
\def\confYear{2026}

\title{CLIP Is Shortsighted: Paying Attention Beyond the First Sentence}

\author{Marc-Antoine Lavoie$^{1}$\thanks{Work done while the author was an intern at Mila. \\ Correspondence: marc-antoine.lavoie@robotics.utias.utoronto.ca \\ Code: \url{https://github.com/TRAILab/DeBias-CLIP.git}} \quad Anas Mahmoud$^2$ \quad Aldo Zaimi$^2$ \\ Arsène Fansi Tchango$^2$ \quad Steven L. Waslander$^1$ \\
$^1$University of Toronto Robotics Institute \quad $^2$Mila - Quebec AI Institute \vspace{-60pt}}

\begin{document}
\maketitle
\begin{strip}
    \centering
    \includegraphics[width=0.88\linewidth]{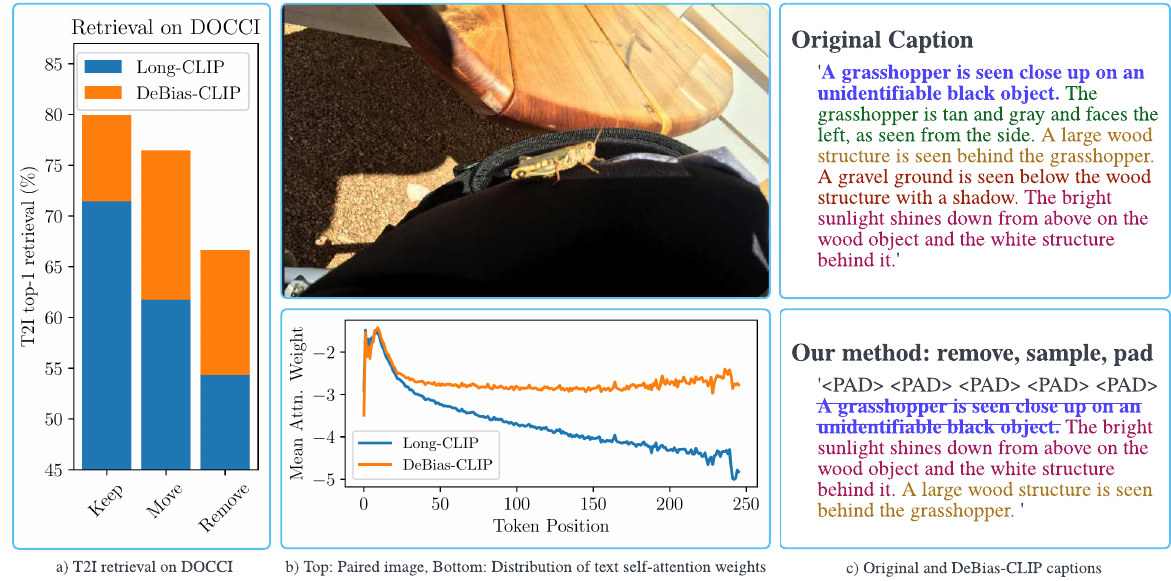}
    \captionof{figure}{\textbf{Key issues in Long-CLIP and our proposed DeBias-CLIP.} a) CLIP models fine-tuned on long captions, such as Long-CLIP~\cite{zhang2024long}, are biased towards early tokens and expect a summary-like first sentence in captions to obtain good retrieval performance. On DOCCI text-to-image retrieval, swapping the first and fourth sentences of the long caption (\emph{Move}) substantially degrades performance ($-9.7\%$), and removing the summary sentence (\emph{Remove}) is even more detrimental ($-17.1\%$). b) Long-CLIP shows a steady decline in self-attention as a function of token depth, while our Debias-CLIP has consistent token attention. c) Our \textbf{DeBias-CLIP} method resolves these issues with three caption-level text augmentations: starting from the original caption, we (i) \textbf{remove} the opening summary sentence, (ii) randomly \textbf{sample} from the remaining sentences, and (iii) \textbf{pad} the tokenized sequence to increase exposure of later positional embeddings during training.}
    \label{fig:placeholder}
\end{strip}


\begin{abstract}
CLIP models learn transferable multi-modal features via image-text contrastive learning on internet-scale data. They are widely used in zero-shot classification, multi-modal retrieval, text-to-image diffusion, and as image encoders in large vision-language models.  However, CLIP’s pretraining is dominated by images paired with short captions, biasing the model toward encoding simple descriptions of salient objects and leading to coarse alignment on complex scenes and dense descriptions. While recent work mitigates this by fine-tuning on small-scale long-caption datasets, we identify an important common bias: both human- and LLM-generated long captions typically begin with a one-sentence summary followed by a detailed description. We show that this acts as a shortcut during training, concentrating attention on the opening sentence and early tokens and weakening alignment over the rest of the caption. To resolve this, we introduce DeBias-CLIP, which removes the summary sentence during training and applies sentence sub-sampling and text token padding to distribute supervision across all token positions. DeBias-CLIP achieves state-of-the-art long-text retrieval, improves short-text retrieval, and is less sensitive to sentence order permutations. It is a drop-in replacement for Long-CLIP with no additional trainable parameters.





\end{abstract}    
\section{Introduction}
\label{sec:intro}


CLIP encoders learn a joint text-image representation space through contrastive learning on large-scale datasets. This alignment not only yields a powerful visual encoder but also facilitates zero-shot transfer for visual tasks by leveraging the corresponding text encoder. Furthermore, the shared representation space serves as a foundation for a broad range of vision-language applications. For instance, text-to-image diffusion models such as Stable Diffusion~\cite{podellsdxl} utilize CLIP’s text encoder for conditional generation, while its image encoder is widely adopted in generative vision–language models (VLMs)~\cite{liu2023visual, lin2023video, lillava, beyer2024paligemma}.

However, CLIP models have significant limitations. The original CLIP training datasets are dominated by short captions, which leads to models that are primarily sensitive to salient objects in images and early-text tokens and limiting performance on more complex downstream tasks. Similarly, the token limit of 77 of the CLIP text encoder corresponds to only about three or four sentences, meaning that paragraph-length texts cannot be fully encoded.

Recently, many works have shown that training CLIP encoders~\cite{zheng2024dreamlip, xiao2025flair, xie2025fg} or VLMs~\cite{chen2024sharegpt4v, deitke2025molmo} with curated long captions instead of short captions can obtain competitive performance on text-image retrieval, dense vision tasks such as zero-shot segmentation, and visual question answering (VQA), while using millions rather than hundreds of millions of images, though compute remains substantial. Alternatively, Long-CLIP~\cite{zhang2024long} addresses token-length limits by stretching positional embeddings and fine-tuning a pretrained CLIP model on a small long-caption dataset, yielding better retrieval on paragraph-length texts.




And yet, further limitations remain. As shown in~\cref{fig:placeholder}, the standard Long-CLIP model's performance is degraded when caption sentences are permuted or when the first sentence of long captions is removed. We find that this behavior is consistent with an early-token bias observed in the original pretrained CLIP variants~\cite{zhang2024long}, and identify that training CLIP models to align image features with the first sentence in long captions is a key issue. In both training and test long-caption datasets, the first sentence typically serves as a summary of the entire text and generally contains a significant portion of the caption information~\cite{peleg2025advancing}. Since we start from pretrained CLIP models that are mostly trained on short texts, the first sentence acts as a shortcut, allowing the models to minimize the contrastive loss without needing to extend the effective context window size. This leads to reduced sensitivity to later tokens, in contradiction to the objective of extending the effective context window.


To address this, we introduce \emph{DeBias-CLIP}, which mitigates the bias by sampling long captions differently. Our method is a simple drop-in replacement for Long-CLIP that does not add any additional trainable parameters. By excluding the opening summary sentence from the sampled caption text and by using sentence sampling and token padding, we obtain state-of-the-art performance on many long retrieval datasets while also improving the performance on short retrieval. Finally, our method also significantly improves the robustness of retrieval with permuted sentences and when the summary sentence is removed.

Our main \textbf{contributions} in this work are the following:
\begin{itemize}
    \item We empirically show that existing CLIP models and models trained with the Long-CLIP framework show a significant bias towards early text tokens and information-dense summary sentences that are common in long-caption datasets, with degraded performance when information is moved deeper in captions.
    \item We propose a simple augmentation strategy to mitigate this bias by removing the summary sentence from the training captions, sub-sampling the remaining caption sentences, and adding padding tokens. Unlike prior work, our method does not introduce new trainable parameters (e.g.,~\cite{xie2025smartclip, asokan2025finelip}) and does not require multi-stage or student-teacher frameworks (e.g.,~\cite{najdenkoska2024tulip}) to adapt CLIP to a long-context window.
    \item Our method achieves state-of-the-art performance on multiple long-text retrieval tasks, in addition to greatly reduced sensitivity to sentence permutation in captions and to the absence of summary sentences. The improvements are consistent across multiple scales and pretrained CLIP models.
\end{itemize}

\section{Related Work}
\label{sec:related_works}

\subsection{Contrastive Vision-Language Pretraining}



Contrastive Vision-Language Pretraining (CLIP)~\cite{radford2021learning} and similar works, such as ALIGN~\cite{jia2021scaling}, have popularized a dual-encoder, contrastive pretraining paradigm in which image and text encoders are trained jointly on large collections of image-caption pairs. The objective maximizes the similarity of matched pairs and minimizes that of mismatches in a mini-batch. This approach has been shown to yield robust representations that support various downstream tasks such as zero-shot classification~\cite{radford2021learning, zhou2022conditional, gao2024clip}, image generation~\cite{rombach2022high, ramesh2022hierarchical}, detection~\cite{gu2021open, li2022grounded, zhong2022regionclip, zhou2022detecting}, segmentation~\cite{li2022language, xu2022groupvit, luddecke2022image}, VQA~\cite{li2023blip, shen2021much, song2022clip}, and video understanding~\cite{luo2022clip4clip, xu2021videoclip}. Subsequent work has focused on refining data, losses, or architectures while keeping the dual-encoder recipe. OpenCLIP~\cite{cherti2023reproducible} scales training to billions of image-text pairs, and SigLIP~\cite{zhai2023sigmoid} replaces softmax with pairwise sigmoid for better scaling. SLIP~\cite{mu2022slip} and SILC~\cite{naeem2024silc} add image-only self-supervised losses, LiT~\cite{zhai2022lit} aligns text features to a frozen image encoder, and SuperCLIP~\cite{zhaosuperclip} predicts pre-encoder text tokens from images. DeCLIP~\cite{li2021supervision} adds image/text self-supervision and cross-modal alignment, CoCa~\cite{yu2022coca} adds caption generation, and SigLIP2~\cite{tschannen2025siglip} combines multiple losses with large-scale training to obtain a stronger baseline. However, these methods do not generally consider performance on longer texts.


\subsection{Extending CLIP to Long Captions}

CLIP models were originally trained with a short context window (77 tokens), which limits their ability to align images with long, multi-sentence texts. Long-CLIP~\cite{zhang2024long} addresses this by linearly interpolating the pretrained model’s positional embeddings to a longer window and fine-tuning with two contrastive objectives: (i) align global image embeddings to long captions and (ii) align primary image components (via PCA) to short-caption segments. The use of stretched positional embeddings has since been adopted in several state-of-the-art long-caption retrieval methods, including FineLIP~\cite{asokan2025finelip} and SmartCLIP~\cite{xie2025smartclip}. Alternatively, TULIP~\cite{najdenkoska2024tulip} swaps interpolation for rotary positional embeddings~\cite{su2024roformer}, which have been shown to scale better in large language models, but it requires two-stage training.

One central challenge for long-caption CLIP models is preserving short-text retrieval while improving performance on long captions. A common approach is to train on both short and long captions, but this requires mechanisms that prevent long-text representations from collapsing to the short-text representations. Long-CLIP does this by using PCA on the image features matched to short captions; Fix-CLIP~\cite{wang2025fix} masks image tokens aligned to short captions and learns local image aggregation tokens; SmartCLIP learns a text-conditional masking network to mask image feature channels that are not significant for alignment (adding more parameters); FLAIR~\cite{xiao2025flair} uses text features as queries for image-feature pooling, which requires all text features at inference and scales poorly. Finally, some methods reuse the original CLIP encoder. LongD-CLIP~\cite{feng2025retaining} trains with a second distillation phase in which a model finetuned on long captions is realigned with the original CLIP model, increasing complexity. In contrast, we aim to minimize extra parameters, encoders, and training stages. We find that pretrained CLIP models suffer from early-token and summary-sentence biases that persist through long-text training, and that simple text augmentations mitigate these biases and improve performance without introducing additional overhead.

\subsection{Vision-Language Datasets}

Large-scale contrastive pretraining has been enabled by web-scale image-text collections with captions such as alt-text or brief descriptions. Examples include CC12M~\cite{changpinyo2021conceptual}, Visual Genome~\cite{krishna2017visual}, SBU~\cite{ordonez2011im2text}, COCO~\cite{lin2014microsoft}, and LAION-5B~\cite{schuhmann2022laion}. Their scale and diversity are crucial for representation learning, but the captions are typically short, formulaic, and noisy, which limits supervision for fine-grained grounding and long-context understanding. 

Recent work shows that contrastive training on curated long captions enables training of VLMs (including CLIP-like models) on much less data than internet-scale short-caption corpora that more traditional methods require. 
ShareGPT4V~\cite{chen2024sharegpt4v} leverages GPT-4V~\cite{openai2023gpt4v} captions to first train a labeler and then recaptions 1.2M images, and works such as Long-CLIP and SmartCLIP show that finetuning pretrained models at that data scale substantially improves long-text retrieval. Scaling further, DreamLIP uses the ShareGPT4V labeler to annotate 30M images, sufficient to train a model from scratch, while at the billion scale, long captions remain advantageous \cite{li2024if}. For evaluation, more curated long-text retrieval benchmarks such as DCI~\cite{urbanek2024picture}, Urban1K~\cite{zhang2024long}, and DOCCI~\cite{onoe2404docci} probe models’ ability to match images with long, information-dense queries, where crucial details may appear later in the caption.


Similar to prior work~\cite{zhang2024long, xie2025smartclip, najdenkoska2024tulip}, we adapt pretrained CLIP models using ShareGPT4V for long-caption supervision and evaluate across both short- and long-text retrieval benchmarks (DCI, Urban1K, DOCCI). Our method targets an information position bias common in dense-caption datasets and improves token-level attention uniformity, yielding strong gains on long-text retrieval.

\section{Empirical Analysis of CLIP Biases}
\label{sec:3_empirical_analysis}
In this section, we present an empirical analysis of how the text encoders of existing CLIP-like vision-language models behave in long-text retrieval tasks. First, we highlight shared elements in long-text retrieval datasets that affect training and evaluation. Next, we use long-text datasets to show that pretrained CLIP models have a limited effective context window that is biased towards both early tokens and overview/summary sentences. Finally, we show how training on long-context data does not resolve these biases. Long-text retrieval models trained using the Long-CLIP recipe show significant loss in performance under simple, representative augmentations such as removing or moving the opening/summary sentence, which are similar to texts used in retrieval on generic chunked text from larger documents and Retrieval-Augmented Generation (RAG) pipelines.



\begin{table*}[th!]
\centering
    {\begin{tabular}{cccc}
    \toprule
        Dataset & Count & Avg. Tokens & Example Caption \\
    \hline
    COCO~\cite{lin2014microsoft}         & 5k      & 13.5
        & ``A black Honda motorcycle parked in front of a garage'' \\
    Flickr30k~\cite{young2014image}    & 30k     & 15.8
        & ``A group of people stand in the back of a truck filled with cotton.'' \\
    \hline 
    & & & \\[-0.8em] 
    \multirow{2}{*}{ShareGTP4V~\cite{chen2024sharegpt4v}}    & \multirow{2}{*}{1.2M}      & \multirow{2}{*}{168.6} & \parbox[t]{10cm}{
    \linespread{0.5} \selectfont
    ``\textbf{This image captures a serene moment at a zoo, featuring three majestic giraffes in their enclosure.} The giraffes, with their distinctive...''} \\
    & & & \\[-0.8em]
    \multirow{2}{*}{Urban1k~\cite{zhang2024long}}      &\multirow{2}{*}{1k}        & \multirow{2}{*}{131.4}
        & \parbox[t]{10cm}{\linespread{0.5} \selectfont ``\textbf{This image captures a lively street market under daylight.} Fresh produce is on display in wooden crates directly in the foreground...''} \\
    & & & \\[-0.8em]
    \multirow{3}{*}{DCI~\cite{urbanek2024picture}}          & \multirow{3}{*}{7.8k}      & \multirow{3}{*}{174.2}
        & \parbox[t]{10cm}{\linespread{0.5} \selectfont ``\textbf{A series of illuminated escalators in an indoor shopping area.}'' \\ ``\textbf{An indoor mall with three illuminated escalators.} The mall has various planters with lush greenery on both sides of the escalator... ''} \\
    & & & \\[-0.8em]
    \multirow{2}{*}{DOCCI~\cite{onoe2404docci}}        & \multirow{2}{*}{5k}        & \multirow{2}{*}{141.5}
        & \parbox[t]{10cm}{\linespread{0.5} \selectfont ``\textbf{A high angle view of an old faded street corner.} In the middle of the view is the orange spray painted word "ROW", with a horizontal... ''} \\
    \bottomrule
    \end{tabular}}
    \caption{\textbf{Short- and long-text retrieval datasets.} For long-caption datasets, the opening summary sentence is \textbf{bolded}. DCI provides both a short caption and a separate long caption~\cite{najdenkoska2024tulip}; in both cases, the short caption and the first sentence of the long caption summarize the image. The token count statistics are calculated using the CLIP tokenizer.}
    \label{tab:retrieval_datasets}
\end{table*}

\subsection{Text Retrieval Datasets}
\label{sec:31_emp_datasets}
Following the Long-CLIP method~\cite{zhang2024long}, we train models on ShareGPT4V~\cite{chen2024sharegpt4v} and evaluate on common long-text retrieval benchmarks: Urban1k~\cite{zhang2024long}, DCI~\cite{urbanek2024picture}, Long-DCI~\cite{najdenkoska2024tulip}, and DOCCI~\cite{onoe2404docci}. In addition, we also evaluate on the COCO~\cite{lin2014microsoft} and Flickr30k~\cite{young2014image} short-text retrieval datasets. \cref{tab:retrieval_datasets} presents summary information on all datasets. 


First, we note that short-text retrieval datasets have caption lengths well below CLIP’s 77-token limit (64 for SigLIP), and so provide no information on long-text performance. 
Next, we highlight how these long-caption datasets share a similar format: they start with a \textbf{summary first sentence} followed by sentences with fine-grained details, and they all have a similar average token length. While common in human- and LLM-generated captions, this structure is atypical in practical image–text retrieval (e.g., chunked passages from larger documents). Because the summary sentence contains much of the caption information, this is similar to training with short captions, and this structure allows biases in pretrained models to persist in long-context models. We present our analysis on DOCCI, but consider additional datasets in the Supplementary Materials.


\subsection{Biases in CLIP Text Encoders}
\label{sec:32_biases_clip}
In this study, we consider VLMs that follow the original \mbox{OpenAI} CLIP (CLIP) model's design, with variations of training data with the OpenCLIP model trained on LAION-2B (OpenCLIP), and differences in both data and loss with Google's SigLIP~\cite{zhai2023sigmoid} and SigLIP2~\cite{tschannen2025siglip} trained on WebLI~\cite{chenpali}. In all cases, the training datasets are dominated by pairs of images and single-sentence captions. Using multi-sentence long-caption datasets, we show that their text encoders exhibit a systematic early-token bias and pronounced sensitivity to summary/overview sentences. 

\begin{figure}[t]
  \centering
    \includegraphics[width=0.47\textwidth]
    {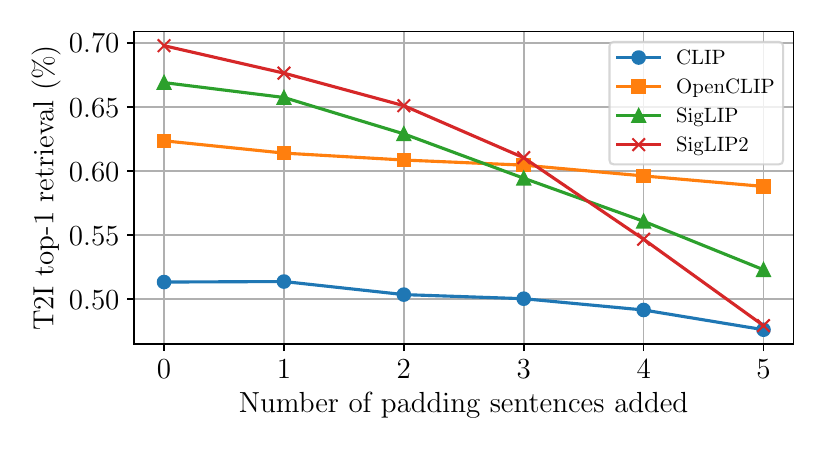}
        \caption{\textbf{Top-1 text-to-image retrieval on DOCCI as a function of the number of added padding sentences.} One to five padding sentences \texttt{`This is a photo.'} are added before the truncated original DOCCI caption (we keep the first two sentences only). We use the ViT-B/16 scale for all models.}
    \label{fig:retrieval_with_padding}
\end{figure}

\subsubsection{Early-Token Bias}
First, we show that CLIP models have a consistent bias towards early tokens. Using the first two sentences of DOCCI captions (average of 46.6 tokens, well within the models’ context windows) for image-to-text retrieval, we evaluate sensitivity to token position by prepending copies of uninformative sentences \texttt{`This is a photo.'}. These padding tokens shift informative tokens later without adding semantic information. \cref{fig:retrieval_with_padding} shows that prepended padding reduces retrieval performance across all models. Notably, this drop is very large for SigLIP and SigLIP2, even with one or two padding sentences. Building on the analysis in Long-CLIP of the effective context window size~\cite{zhang2024long}, we show that the lack of improvement as the number of tokens increases is at least in part caused by the inability to effectively use information in later token positions rather than purely because of context length. Moving informative content later in the sequence hurts retrieval across CLIP variants.


\begin{figure}[t]
  \centering
    \includegraphics[width=0.46\textwidth]
    {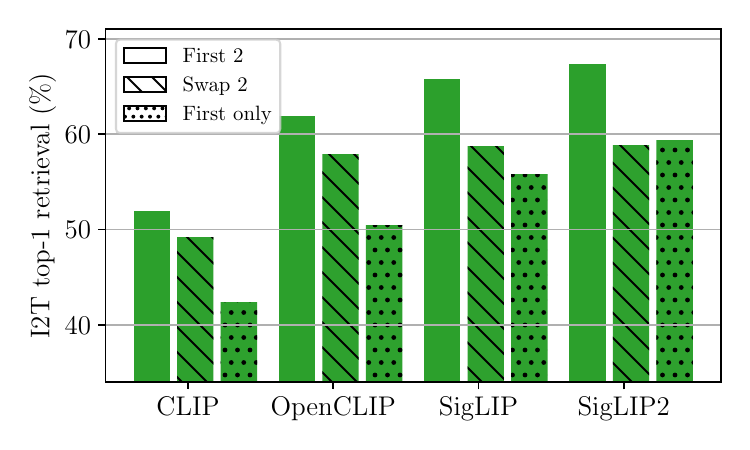}
        \caption{\textbf{Top-1 image-to-text retrieval on DOCCI with first two sentences permuted.} We analyze three setups: the first two sentences in the correct order (\emph{First 2}), the same two sentences swapped (\emph{Swap 2}), and the first sentence only (\emph{First only}). Results are reported for four models: OpenAI CLIP, OpenCLIP (LAION-2B), SigLIP, and SigLIP2.}
    \label{fig:retrieval_with_swap}
\end{figure}

\subsubsection{Sensitivity to Sentence Permutation}
We further evaluate the token position sensitivity of CLIP text encoders by swapping the first two sentences of DOCCI captions. We evaluate retrieval for three cases: (i) the first two sentences in the correct order (\emph{First 2}), (ii) the same two sentences swapped (\emph{Swap 2}), and (iii) the first sentence only (\emph{First only}). \cref{fig:retrieval_with_swap} presents the results and shows that swapping the first two sentences (pushing the summary sentence later in the sequence) reduces performance for all models. SigLIP2 drops the most at $-8.5\%$. For both correctly ordered and single sentence cases, models that are better at short retrieval~\cite{tschannen2025siglip} are also better on DOCCI, but this is not the case under permutation, with larger drops for SigLIP and SigLIP2. This shows that SigLIP models are particularly sensitive to token position, and here the improvements from the second sentence's added information are nullified if the summary sentence is pushed later in the context window.

\subsection{Effects on Long-Text CLIP Variants}
\label{sec:33_biases_longclip}
Finally, we evaluate if the CLIP biases are also present in long-context models and use Long-CLIP as a representative model. We evaluate different cases for \emph{full long-caption retrieval} in \cref{fig:placeholder}. Similar to original CLIP models, Long-CLIP remains sensitive to sentence position: swapping the first (summary) and the fourth sentences (\emph{Keep vs Move}) leads to a $-9.7\%$ drop, and removing the first sentence altogether (\emph{Keep vs Remove}) drops by $-17.1\%$. These results indicate that (i) the model struggles to use summary information when it is not in the first few tokens, and (ii) accurate retrieval still requires the summary sentence. We see a similar trend when considering the averaged pre-softmax attention weights in \cref{fig:placeholder}, which aggregate the output token's attention towards the other text token positions at the transformer's last layer. For Long-CLIP, attention peaks on early tokens and decays steadily beyond the first 10 positions. The sums are normalized by the number of times a token position is within the text window (i.e., not post-padding), which explains the higher variance at later tokens. 
\section{Method}
\label{sec:4_method}

\begin{table}[t]
\centering
\small{
\begin{tabular}{|c||c|c|c|c|c|c|c|}
\hline 
\cellcolor{Turquoise!40} \textbf{Original} & \multirow{2}{*}{\texttt{SOT}} & \multirow{2}{*}{$\mathtt{s}_{1}$} & \multirow{2}{*}{$\mathtt{s}_{2}$} & \multirow{2}{*}{$\mathtt{s}_{3}$} & \multirow{2}{*}{$\mathtt{s}_{4}$} & \multirow{2}{*}{\texttt{EOT}} & \multirow{2}{*}{\texttt{PAD}} \\
\cellcolor{Turquoise!40} \textbf{Caption} & & & & & & & \\
\hline
\hline
\cellcolor{Tan!40} \textbf{Our} & \multirow{2}{*}{\texttt{SOT}} & \multirow{2}{*}{\texttt{PAD}} & \multirow{2}{*}{\texttt{PAD}} & \multirow{2}{*}{$\mathtt{s}_{4}$} & \multirow{2}{*}{$\mathtt{s}_{2}$} & \multirow{2}{*}{\texttt{EOT}} & \multirow{2}{*}{\texttt{PAD}} \\
\cellcolor{Tan!40} \textbf{Caption} & & & & & & & \\
\hline
\end{tabular}
}
\caption{\textbf{Visualization of our caption augmentation method}. We drop the summary sentence $\mathtt{s}_{1}$, randomly sample from the other sentences, and add padding (\texttt{PAD}) before the first sampled sentence. We do not enforce the preservation of sentence order when sampling. Here, \texttt{SOT} and \texttt{EOT} refer to start-of-text and end-of-text tokens, respectively.
}
\label{tab:our_caption}
\end{table}

In this section, we present three caption-level training augmentations to mitigate the summary sentence shortcut: \textbf{dropping} the summary sentence, randomly \textbf{sampling} from the remaining long-caption sentences, and \textbf{padding} the beginning of the text sequence. Our augmentations are summarized in \cref{tab:our_caption}, and we otherwise follow the Long-CLIP~\cite{zhang2024long} method. 

\subsection{Problem Definition}
For our analysis and proposed sampling strategy, we treat captions as a set of concatenated sentences, each contributing meaningful semantic concepts. Although cross-sentence context can be beneficial, we approximate sentences as independent semantic units for sampling purposes. Given a long caption $C$ with $k$ sentences ${n_{\mrm{sents}} = k}$, we model it as ${C = [s_{1},\,s_{2},\,\dots,\,s_{k}]}$.
The caption $C$ can be tokenized with a tokenizer $\varphi$ as ${T = \varphi \left( C \right)}$, giving ${T =[\mathtt{SOT},\,\mathtt{t}_{1},\,\mathtt{t}_{2},\,\dots,\,\mathtt{t}_{k},\,\mathtt{EOT},\,\mathtt{PAD}]}$, where $\mathtt{SOT}$, $\mathtt{EOT}$ and $\mathtt{PAD}$ are respectively the start-of-text, end-of-text and padding tokens. For Long-CLIP, the short and long training captions are given by
\begin{align}
    C_{LC}^{s} &= [s_{1}], \\
    C_{LC}^{\ell} &= [s_{1},\,s_{2},\,\dots,\,s_{k}].
\end{align}
Long-CLIP encodes short and long captions, and images as feature vectors $u^{s}$, $u^{\ell}$, and $v$ respectively, and the contrastive loss $\mathcal{L}^{\mrm{cont}}$ is evaluated independently for both captions:
\begin{align}
    \mathcal{L}^{\mrm{cont}}\left(a,b\right) &= -\dfrac{1}{N} \sum_{i=1}^N \log\dfrac{\exp \left(\mrm{sim}\left(a_i,b_i\right) / \tau\right)}{\sum_{j=1}^N \exp \left(\mrm{sim}\left(a_i,b_j\right) / \tau\right)}, \nonumber \\
    \mathcal{L}^{s} &= \mathcal{L}^{\mrm{cont}}\left(u^{s},f\left(v\right)\right) + \mathcal{L}^{\mrm{cont}}\left(f\left(v\right),u^{s}\right), \label{eq:short_cap_cl}   \\
    \mathcal{L}^{\ell} &= \mathcal{L}^{\mrm{cont}}\left(u^{\ell},v\right) + \mathcal{L}^{\mrm{cont}}\left(v,u^{\ell}\right), \label{eq:long_cap_cl} \\
    \mathcal{L} &= \mathcal{L}^{s} + \mathcal{L}^{\ell},
\end{align}
where $\mrm{sim}$, $\tau$, $N$, and $f\left(\cdot\right)$ are respectively a similarity function, a temperature parameter, the number of pairs in a batch, and a batch-wise PCA operator~\cite{zhang2024long} that approximates the image feature vector from a low-rank decomposition.

\subsection{Replacing the Summary Sentence}
As discussed in \cref{sec:31_emp_datasets}, ShareGPT4V captions start with a summary sentence. Long-CLIP uses this first sentence as a short caption to preserve short-text performance. Our short caption serves a different purpose: it is used to capture fine-grained details that may be ignored in the full caption (due to appearing later). We preserve short-text performance via subsampling and prefix padding, not by relying on the summary sentence (see \cref{sec:43_sent_sampling} and \cref{sec:44_token_padding}).

We claim that existing CLIP and Long-CLIP models are sensitive to the inclusion and position of the summary caption, and that, in practice, these sentences can act as a shortcut for the alignment loss when training on captions that always contain this summary. In fact, when evaluating Long-CLIP on the DOCCI retrieval task, we find that the average image-caption similarities for positive pairs are larger when using the summary caption only {$\left(\overline{\mrm{sim}}\left(u^{s},v\right) = 0.320\right)$} compared to using the full caption {$\left(\overline{\mrm{sim}}\left(u^{\ell},v\right) = 0.308\right)$}.

Instead of taking the summary sentence as the short caption, we define the short caption as \emph{the long caption minus its first sentence}, encouraging the model to be sensitive to fine-grained details deeper in the caption. Note that we still use the same long caption $C^{\ell}$ as Long-CLIP to train on the complete text data, and add a short caption given by 
\begin{align}
    C^{\mrm{no\_sum}} &= [s_{2},\,\dots,\,s_{k}].
\end{align}

\subsection{Sentence Sampling}
\label{sec:43_sent_sampling}
To increase the difference between the long caption and the new caption $C^{\mrm{no\_sum}}$ and to encourage the model to be more sensitive to details throughout texts and images, we subsample the sentences in $C^{\mrm{no\_sum}}$ to generate a new short caption. This introduces variation at minimal cost, in contrast to rewriting-based methods used in prior works (e.g., FG-CLIP~\cite{xie2025fg}, LaCLIP~\cite{fan2023improving}, and TIPS~\cite{maninis2024tips}). We randomly sample without replacement ${n_{\mrm{sampled}} = \mathcal{U}\left\{1\,,2\,,\dots,\,n_{\mrm{sents}}-1\right\}}$ sentences in $C^{\mrm{no\_sum}}$ (excludes the first sentence), leading to captions with a wide range of lengths. We do not maintain sentence order when sampling. A possible sampled short caption is given by
\begin{align}
    C^{\mrm{samp}} &= [s_{4},\,s_{2}]. \label{eq:caption_sampling}
\end{align}

\subsection{Token Padding}
\label{sec:44_token_padding}
When adding sentence sampling, performance does not always improve and sometimes gets worse. We hypothesize that this could be related to uneven training of the positional embeddings, as the long caption $C^{\ell}$ can be biased towards the summary sentence and thus early embeddings, in addition to the sampled caption $C^{\mrm{samp}}$ being shorter. Moreover, the pretrained model's short-text performance is better preserved when training with the summary sentences because those captions are more similar to the original training data, but this is no longer the case with the sampled captions. To resolve both issues, we add padding tokens to the start of the tokenized sampled caption to push it to a further position. In encoders with a \texttt{SOT} token (such as CLIP), this padding is added after the \texttt{SOT} token. Given $n_{\mrm{post}}$ post-caption padding tokens, we transfer $n_{\mrm{pre}}$ of them as pre-caption padding tokens and update the post-padding as
\begin{align}
    n_{\mrm{pre}} &= \mathcal{U}\left\{0,\,1,\,\dots,\,n_{\mrm{post}}\right\}, \\
    n_{\mrm{post}} &= n_{\mrm{post}} - n_{\mrm{pre}}.
\end{align}
This reassigns part of the post-caption padding to the prefix \emph{without truncating any text tokens}. Given the sampled caption from \eqref{eq:caption_sampling}, our final tokenized short caption would be
\begin{align}
    T^{s}_{\mrm{ours}} &= [\mathtt{SOT},\,\mathtt{PAD}_{\mrm{pre}},\,\mathtt{s}_{4},\,\mathtt{s}_{2},\,\mathtt{EOT},\,\mathtt{PAD}_{\mrm{post}}]. \label{eq:caption_padding}
\end{align}

\subsection{Multi-caption Training}
Given the tokenized long caption ${T^{\ell}_{LC} = T^{\ell}_{\mrm{ours}}}$ and the sampled and padded caption $T^{s}_{\mrm{ours}}$ from \eqref{eq:caption_padding}, we compute independent contrastive losses for each caption with \eqref{eq:short_cap_cl} and \eqref{eq:long_cap_cl} and consider the weighted sum of both as the final loss
\begin{align}
    \mathcal{L} &= \lambda^{s} \mathcal{L}^{s} + \left(1-\lambda^{s}\right) \mathcal{L}^{\ell}, \label{eq:weighted_loss}
\end{align}
where $\lambda^{s}$ is a weight coefficient that controls the relative contribution of the short and long caption losses.


\section{Experiments and Results}
\label{sec:5_experiments}

\begin{table*}[ht!]
\centering
\begin{tabular}{c|c c cc cc cc cc}
\toprule \multirow{2}{*}{} & \multirow{2}{*}{Method} & \multirow{2}{*}{Venue} & \multicolumn{2}{c}{Urban1k} & \multicolumn{2}{c}{DCI} & \multicolumn{2}{c}{Long-DCI} & \multicolumn{2}{c}{DOCCI} \\
& & &  T2I & I2T & T2I & I2T & T2I & I2T & T2I & I2T \\
\midrule
\parbox[t]{2.5mm}{\multirow{8}{*}{\rotatebox[origin=c]{90}{ViT-B/16}}}
& CLIP~\cite{radford2021learning}            & ICML21 & 53.4 & 67.5 & 42.9 & 44.1 & 32.7 & 35.9 & 57.1 & 60.6 \\
& Long-CLIP~\cite{zhang2024long}      & ECCV24 & 79.5 & 78.9 & 57.1 & 51.6 & 47.0 & 41.1 & 71.4 & 63.1 \\
& TULIP~\cite{najdenkoska2024tulip}          & ICLR25 & 86.6 & 88.1 & -    & -    & 50.6 & 50.2 & -    & -    \\
& FineLIP~\cite{asokan2025finelip}       & CVPR25 & \underline{90.0} & \underline{91.2} & -    & -    & -    & -    & \underline{78.1} & \textbf{80.0} \\
& LongD-CLIP~\cite{feng2025retaining}     & CVPR25 & 87.3 & 87.2 & -    & -    & -    & -    & -    & -    \\
& SmartCLIP~\cite{xie2025smartclip}      & CVPR25 & 87.4 & 90.0 & \underline{64.0} & \underline{64.9} & \underline{52.8} & \underline{53.4} & 78.0 & 77.4 \\
& Fix-CLIP~\cite{wang2025fix}  & ICCV25 & 81.1 & 80.9 & 63.0 & 59.7 & -    & -    & -    & -    \\
& \textbf{DeBias-CLIP (ours)}        & -      & \textbf{93.0} & \textbf{93.1} & \textbf{67.6} & \textbf{68.5} & \textbf{57.4} & \textbf{57.8} & \textbf{80.0} & \underline{79.7} \\
\midrule
\parbox[t]{2.5mm}{\multirow{8}{*}{\rotatebox[origin=c]{90}{ViT-L/14}}}
& CLIP            & ICML21 & 56.1 & 68.5 & 43.8 & 44.8 & 33.9 & 36.0 & 63.0 & 65.8 \\
& Long-CLIP       & ECCV24 & 86.0 & 82.5 & 63.9 & 57.0 & 52.7 & 45.5    & 78.6   & 66.5 \\
& TULIP           & ICLR25 & 91.1 & 90.1 & 66.2 & 66.0 & 56.4 & 55.7 & 79.1 & 77.9 \\
& FineLIP         & CVPR25 & \underline{93.9} & \underline{94.5} & -    & -    & \underline{60.7} & \underline{60.8} & \underline{84.4} & \underline{83.7} \\
& LongD-CLIP      & CVPR25 & 90.8 & 91.9 & -    & -    & -    & -    & -    & -    \\
& SmartCLIP       & CVPR25 & 90.1 & 93.3 & \underline{69.8} & \underline{68.2} & 58.5 & 57.6 & 82.5 & 81.6 \\
& Fix-CLIP  & ICCV25 & 87.7 & 86.8 & 66.7 & 65.1 & -    & -    & -    & -    \\
& \textbf{DeBias-CLIP (ours)}  & -      & \textbf{95.2} & \textbf{95.2} & \textbf{73.5} & \textbf{72.8} & \textbf{63.4} & \textbf{62.4} & \textbf{85.6} & \textbf{85.2} \\
\bottomrule
\end{tabular}
\caption{\textbf{Comparison of top-1 long-text retrieval on standard benchmarks}. Best performing methods are \textbf{bolded} and second best are \underline{underlined}. We report numbers from the original papers when available and otherwise evaluate using our implementation when checkpoints are provided (Long-CLIP, TULIP VIT-L, SmartCLIP).
}
\label{tab:res_long_retrieval}
\end{table*}

\subsection{Datasets}
We train on ShareGPT4V~\cite{chen2024sharegpt4v} and evaluate on the following datasets: Urban1k~\cite{zhang2024long}, DCI~\cite{urbanek2024picture}, and DOCCI~\cite{onoe2404docci}. We summarized dataset information in \cref{tab:retrieval_datasets}. In the tables, DCI is evaluated with the short and long captions concatenated, while Long-DCI~\cite{najdenkoska2024tulip} is a variation that uses only the long caption. We also evaluate short-text retrieval on the 5k validation split of COCO2017 and on the full Flickr30k dataset to confirm that training for long retrieval does not negatively affect performance on short texts.

\subsection{Experimental Setup}
Following Long-CLIP, we start from a pretrained CLIP encoder (unless specified, the OpenAI weights). The text encoder's positional embeddings are extended from 77 to 248 tokens, freezing the first 20 embeddings and using linear interpolation to stretch the rest by a factor of 4. We train for 3 epochs with a batch size of 256 on 4 A100 GPUs. We present results for 3 epochs in the main table to compare directly with SmartCLIP, which performs well on short and long retrieval and adds only a small masking module (not used at inference). Prior work trains with different numbers of epochs, e.g., 1 for Long-CLIP and 6 for FineLIP.

\subsection{Results}
\label{sec:53_results}

\subsubsection{Long-Text Retrieval}
We compare DeBias-CLIP against other state-of-the-art long-text retrieval methods trained on ShareGPT4V in \cref{tab:res_long_retrieval}. Our method exceeds prior work on all datasets except for ViT-B/16 image-to-text retrieval on DOCCI, where we are roughly on par with FineLIP. We perform better on ViT-L/14 by $+1.5\%$, and note that FineLIP adds a cross-modal feature refinement module used at inference (i.e., requires knowing true positive pairs a priori). Relative to SmartCLIP, our closest setup, we see consistent gains across datasets, with especially large improvements on Urban1k T2I ($+5.6\%$ and $+5.1\%$ for ViT-B and ViT-L, respectively). SmartCLIP trains with a single contrastive loss and samples sentences \emph{starting with the summary sentence}. In contrast, DeBias-CLIP \emph{removes the opening summary sentence} in the short caption and uses padding to extend attention to later tokens. We find that training without the summary caption is critical for long-text retrieval. Compared to the original Long-CLIP, we improve performance by at least $+10\%$ on most tasks, indicating that our sampling strategy yields substantial gains even with the same architecture.

\begin{table}[t!]
\centering
\begin{tabular}{c|ccccc}
\toprule
\multirow{2}{*}{} & \multirow{2}{*}{Method} & \multicolumn{2}{c}{COCO} & \multicolumn{2}{c}{Flickr30k} \\
& &  T2I & I2T & T2I & I2T \\
\midrule
\parbox[t]{2.5mm}{\multirow{7}{*}{\rotatebox[origin=c]{90}{ViT-B/16}}}
& CLIP~\cite{radford2021learning}           & 32.7 & 51.8 & 24.7 & 44.1 \\
& Long-CLIP~\cite{zhang2024long}            & 40.4 & 57.6 & 34.1 & 46.8 \\
& TULIP~\cite{najdenkoska2024tulip}         & 40.7 & 56.8 & 35.2 & 46.1 \\
& FineLIP~\cite{asokan2025finelip}          & 40.4 & 58.7 & 34.1 & 52.8 \\
& LongD-CLIP~\cite{feng2025retaining}       & 41.1 & 59.4 & 34.5 & 52.9 \\
& SmartCLIP~\cite{xie2025smartclip}         & \underline{42.4} & \textbf{61.9} & \underline{36.3} & \underline{55.6} \\
& \textbf{DeBias-CLIP (ours)}               & \textbf{43.0} & \underline{61.3} & \textbf{36.6} & \textbf{56.6} \\
\midrule
\parbox[t]{2.5mm}{\multirow{7}{*}{\rotatebox[origin=c]{90}{ViT-L/14}}}
& CLIP            & 35.4 & 56.1 & 28.0 & 48.5 \\
& Long-CLIP       & 46.3 & 62.8 & 41.2 & 53.4 \\
& TULIP           & 46.1 & 62.6 & 41.6 & 56.7 \\
& FineLIP         & 46.2 & 63.4 & 42.4 & 62.2 \\
& LongD-CLIP      & 47.0 & 64.4 & 42.2 & 60.2 \\
& SmartCLIP       & \textbf{48.5} & \textbf{66.0} & \underline{43.8} & \underline{63.9} \\
& \textbf{DeBias-CLIP (ours)}   & \underline{48.1} & \underline{65.1} & \textbf{43.9} & \textbf{64.8} \\
\bottomrule
\end{tabular}
\caption{\textbf{Comparison of top-1 short-text retrieval methods on standard benchmarks}. Best performing methods are \textbf{bolded}, second best are \underline{underlined}. We present reported numbers when available and evaluate on other datasets with provided checkpoints with our implementation.
}
\label{tab:res_short_retrieval}
\end{table}

\subsubsection{Short-Text Retrieval}
Similar to our results on long retrieval, we observe consistent gains over the Long-CLIP baseline ($+9.8\%$ and $+11.4\%$ on Flickr I2T for ViT-B and ViT-L, respectively), and we outperform most competing methods. On COCO, however, SmartCLIP is often slightly ahead despite our stronger results on Flickr30k. We show that there can be tradeoffs between short and long retrieval performance in our ablations (see \cref{sec:54_ablations}), and training on full long captions likely emphasizes long-text performance.

\subsubsection{Generalization to other CLIP-Style Encoders}
\label{sec:533_gen_to_other_CLIP}
\begin{figure}[t]
  \centering
    \includegraphics[width=0.45\textwidth]
    {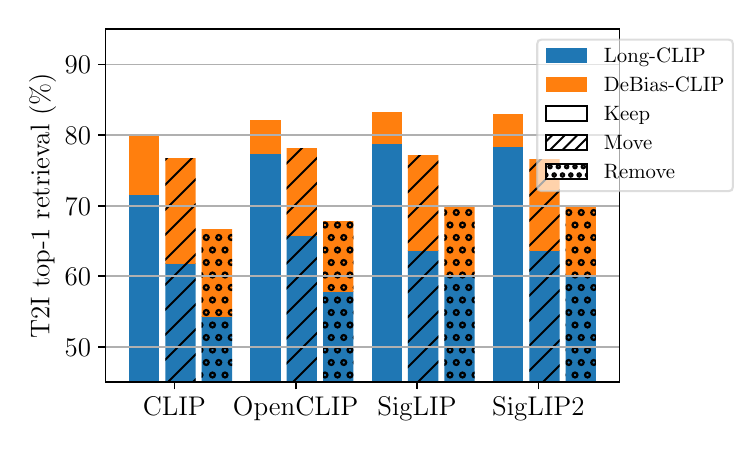}
        \caption{\textbf{Top-1 text-to-image retrieval on DOCCI for different CLIP pretrained models.} We consider 3 cases: \emph{Keep} (full caption), \emph{Move} (swap the first and fourth sentences), and \emph{Remove} (drop the first sentence). We improve performance for all encoders.}
    \label{fig:longclip_mult_encoders}
\end{figure}

All prior works discussed in this section use OpenAI's pretrained CLIP weights. Because newer encoders improve on the original CLIP~\cite{zhai2023sigmoid, tschannen2025siglip}, we evaluate the performance of these models with the Long-CLIP method and ours. We reproduce Long-CLIP's training and apply the same positional embedding extension scheme to all models: freeze the first 20 positions and stretch the remaining positions by a factor of 4 via linear interpolation (we find that stretching all 64 SigLIP positional embeddings degrades short-text retrieval and include results in the Supplementary). 
As shown with the \emph{Keep} bars in \cref{fig:longclip_mult_encoders}, DeBias-CLIP consistently outperforms the original Long-CLIP across pretraining data distributions (OpenAI, LAION-2B), various pretraining losses, and the use of text causal masks (CLIP vs SigLIP/SigLIP2). 

\subsubsection{Robustness to Text Permutations}
\label{sec:534_sentence_permutation}
We previously showed (see \cref{sec:3_empirical_analysis}) that both CLIP and Long-CLIP favor early tokens and rely on the summary sentence. We repeat the analysis of \cref{sec:33_biases_longclip} with additional pretrained CLIP models and find in \cref{fig:longclip_mult_encoders} that DeBias-CLIP narrows the gap between the original and permuted captions on retrieval (\emph{Keep} vs \emph{Move}, Long-CLIP $-9.7\%$, ours $-3.5\%$ for OpenAI CLIP), indicating better use of information regardless of position. However, the reduction is still significant on SigLIP and SigLIP2 ($-6.1\%$ and $-6.5\%$ respectively), suggesting residual position sensitivity from pretraining. We also observe substantial gains in retrieval when the summary sentence is absent (\emph{Remove}), indicating improved matching for images paired with dense descriptions. Finally, \cref{fig:placeholder} shows that DeBias-CLIP maintains a much flatter average attention weight across text tokens compared to Long-CLIP, helping it leverage information deeper in captions.


\subsection{Ablations}
\label{sec:54_ablations}

\subsubsection{Contribution of Caption-Level Augmentations}
Starting from Long-CLIP, we add three caption-level augmentations: (i) replace the summary sentence with the long caption \emph{minus} its summary caption as the short caption, (ii) randomly sample sentences from the remaining set, and (iii) prepend padding tokens to the short caption. We present ablations of our method in \cref{tab:abl_components}. Replacing the first summary sentence is critical for long retrieval, while combining subsampling and padding improves short retrieval by about $+1\%$ with mixed effects on long retrieval. Overall, the full DeBias-CLIP configuration consistently outperforms Long-CLIP on both short and long retrieval.

\begin{table}[t]
\centering
\resizebox{0.47\textwidth}{!}
{\begin{tabular}{lcccc}
\toprule \multirow{2}{*}{Method} & COCO & Flickr & Urban1k & DOCCI \\
 & T2I & T2I & T2I & T2I \\
\midrule
Long-CLIP  & 40.4 & 34.1 & 79.5 & 71.4 \\
\quad + 3 epochs         & 40.6 & 33.9 & 82.7 & 75.2 \\
\quad + No summary       & 42.2 & 35.7 & 92.6 & \textbf{80.9} \\ 
\quad + Subsampling & 41.9 & 35.8 & 92.5 & 80.8 \\ 
\quad + Padding  & \textbf{43.0} & \textbf{36.6} & \textbf{93.0} & 79.7 \\
\bottomrule
\end{tabular}}
\caption{\textbf{Ablation on DeBias-CLIP components}. Best performing methods are \textbf{bolded}. Training with a caption that excludes the summary sentence improves long retrieval (Urban1k and DOCCI). Adding sentence sampling and padding boosts short-caption retrieval (COCO and Flickr) while preserving long-caption gains.
}
\label{tab:abl_components}
\end{table}



\begin{figure}[t]
  \centering
    \includegraphics[width=0.46\textwidth]
    {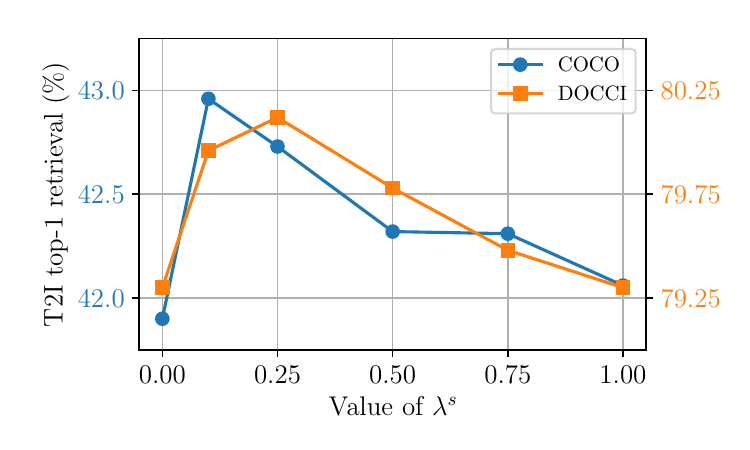}
        \caption{\textbf{Top-1 text-to-image retrieval on COCO (short) and DOCCI (long) as a function of short caption loss weight $\lambda^{s}$.} Short retrieval peaks at $\lambda^{s}=0.1$, while long retrieval peaks at $\lambda^{s}=0.25$. Higher values degrade performance in both cases.}
    \label{fig:effects_lambda}
\end{figure}

\subsubsection{Short-Caption Loss Coefficient} We vary the short-caption loss coefficient from \eqref{eq:weighted_loss} from 0 to 1, and compare performance on COCO and DOCCI in \cref{fig:effects_lambda}. We find that a small weight of ${\lambda^{s} = 0.1}$ provides the best tradeoff between short and long retrieval, while above ${\lambda^{s} = 0.25}$, performance degrades for both settings. 

\section{Conclusion and Limitations}

In this work, we empirically show that CLIP-style text encoders exhibit early-token and summary-sentence biases that hinder long-text retrieval. Standard training approaches maintain these biases because the first sentence in training captions is typically a global summary of the entire long caption. To counter this, we propose \emph{DeBias-CLIP}, which trains without the first summary sentence, uses sentence sampling, and adds text token padding. Our method delivers state-of-the-art long-text retrieval while maintaining short-text performance and improving robustness to sentence permutations and summary sentence removal. However, many long-text benchmarks share the same “summary-first” caption structure, which can mask position bias. Our permutation tests reveal this issue and underscore the need for new evaluation benchmarks that more accurately reflect real-world long-context retrieval, beyond caption datasets (e.g., matching images to multi-paragraph document pages, such as textbooks and reports).

\section*{Acknowledgments}
This research was enabled by compute resources provided by Mila (mila.quebec).


%


{
    \small
    \bibliographystyle{ieeenat_fullname}
    \bibliography{main}
}
\clearpage
\maketitlesupplementary
\appendix

\appendix
\section*{Appendices}
\label{sec:appendix}
\renewcommand{\thesubsection}{\Alph{subsection}}

In \cref{sec:suppl_training_details}, we present the training details of our main results. In \cref{sec:suppl_pseudocode}, we present the pseudocode of our DeBias-CLIP method. In \cref{sec:suppl_additional_ablations}, we present additional ablations, including on the embedding stretching for SigLIP, training length, sampling, padding, and token padding methods. In \cref{sec:suppl_additional_clip_biases} and \cref{sec:suppl_additional_longclip_biases}, we present additional discussion and results on CLIP and Long-CLIP biases, respectively. In \cref{sec:suppl_visualizations}, we present visualizations that highlight the superior long-text retrieval capabilities of DeBias-CLIP and demonstrate its advantages in text-to-image diffusion. 

\subsection{Training Details}
\label{sec:suppl_training_details}
Complete training parameters are provided in \cref{tab:app_training_params}. Our setup largely follows Long-CLIP \cite{zhang2024long}, with two modifications: we train for 3 epochs instead of only 1 and use a smaller batch size of 256 instead of 1024 to ensure consistency across all experimental runs. Experiments were conducted on a variety of hardware configurations, subject to compute cluster availability, but the main results in Sec.~\ref{sec:53_results} of the paper were all run on 4 A100 GPUs.

\begin{table}[h!]
\centering
{\begin{tabular}{cc}
\toprule
Parameter & Value \\
\midrule
Batch size & 256 \\
Training epochs & 3 \\
Warm-up iterations & 200 \\
Weight decay & 1e-2 \\
Learning rate & 1e-6 \\
AdamW $\beta_1$ & 0.9 \\
AdamW $\beta_2$ & 0.999 \\
AdamW $\epsilon$ & 1e-8 \\
\bottomrule
\end{tabular}}
\caption{\textbf{Training parameters}. We use AdamW as the optimizer and run on 4 A100 GPUs for the main paper results.
}
\label{tab:app_training_params}
\end{table}

\subsection{Pseudocode}
\label{sec:suppl_pseudocode}
We present pseudocode of our sampling method in \cref{alg:debias_clip}, where $\varphi\left(\cdot\right)$ is the text tokenizer, $\psi_{\mrm{text}}\left(\cdot\right)$ and $\psi_{\mrm{img}}\left(\cdot\right)$ are respectively the text and image encoders, and $f\left(\cdot\right)$ is an operator that reconstructs feature vectors from their principal components (PCA).  

\begin{algorithm}[t]
\caption{DeBias-CLIP Caption Sampling}
\label{alg:debias_clip}
\begin{flushleft}
\begin{algorithmic}[1]
\Require Long-text caption $C = [s_{1}, s_{2}, \dots, s_{n_{\text{sents}}}]$, image $I$, loss weight $\lambda^{s}$
\Ensure Training loss $\mathcal{L}$
\vspace{0.5em}
\State $C^{\ell} \gets C$ \Comment{Long caption}
\State $T^{\ell} \gets \varphi(C^{\ell})$ \hfill \Comment{Tokenize long caption}
\vspace{0.5em}
\State $C^{\text{no\_sum}} \gets [s_{2}, \dots, s_{n_{\text{sents}}}]$ \Comment{Remove first (summary) sentence}
\State Sample $n_{\text{samp}} \sim \mathcal{U}\{1, \dots, n_{\text{sents}} - 1\}$
\State Sample a subset $\mathcal{S} \subset \{2, \dots, n_{\text{sents}}\}$ with $|\mathcal{S}| = n_{\text{samp}}$ (without replacement)
\State $C^{\text{samp}} \gets$ concatenate $\{s_i : i \in \mathcal{S}\}$
\vspace{0.5em}
\State $T^{\text{samp}} \gets \varphi(C^{\text{samp}})$ \Comment{Tokenize short caption}
\State $T^{\text{samp}} \gets [\mathtt{SOT}, t_1, \dots, t_{n_{samp}}, \mathtt{EOT}, \mathtt{PAD}_{\text{post}}]$
\State Let $n_{\text{post}}$ be the number of post-padding tokens in the tokenized short caption $T^{\text{samp}}$
\State Sample $n_{\text{pre}} \sim \mathcal{U}\{0, \dots, n_{\text{post}}\}$
\State $n_{\text{post}} \gets n_{\text{post}} - n_{\text{pre}}$
\State $T^{s} \gets [\mathtt{SOT}, \underbrace{\mathtt{PAD}, \dots, \mathtt{PAD}}_{n_{\text{pre}}}, t_1, \dots, t_{n_{samp}}, \mathtt{EOT}, \underbrace{\mathtt{PAD}, \dots, \mathtt{PAD}}_{n_{\text{post}}}]$
\vspace{0.5em}
\State Encode captions and image:
\Statex \quad $u^{s} \gets \psi_{\text{text}}(T^{s})$ \hfill \Comment{Short (sampled) caption}
\Statex \quad $u^{\ell} \gets \psi_{\text{text}}(T^{\ell})$ \hfill \Comment{Long caption}
\Statex \quad $v \gets \psi_{\text{img}}(I)$ \hfill \Comment{Image}
\vspace{0.5em}
\State $\mathcal{L}^{s} \gets \mathcal{L}^{s}(u^{s}, f(v))$ \Comment{Short caption loss}
\State $\mathcal{L}^{\ell} \gets \mathcal{L}^{\ell}(u^{\ell}, v)$ \Comment{Long caption loss}
\State $\mathcal{L} \gets \lambda^{s} \,\mathcal{L}^{s} + (1 - \lambda^{s})\,\mathcal{L}^{\ell}$
\State \Return $\mathcal{L}$
\end{algorithmic}
\end{flushleft}
\end{algorithm}

\subsection{Additional Ablations}
\label{sec:suppl_additional_ablations}

\subsubsection{Positional Embedding Stretching for SigLIP}
For the different pretrained CLIP-style models presented in our main results, we follow Long-CLIP \cite{zhang2024long} and stretch the positional embeddings by freezing the first 20 text positions and using linear interpolation to extend the rest by a factor of 4. Results from Long-CLIP justify this for the OpenAI weights, and OpenCLIP weights are trained with the same approach. However, SigLIP models use a different loss, tokenizer, and attention mechanism (with no text causal masking), and therefore, it may not be necessarily appropriate to freeze the first few tokens. We present in \cref{tab:app_abl_stretch_siglip} short and long retrieval results for our DeBias-CLIP method on SigLIP and SigLIP2 weights, comparing stretching all position embeddings against keeping the first 20 frozen. We find that stretching all embeddings leads to a significant reduction in short retrieval ($-15.3\%$ for Flickr T2I with SigLIP2), so we recommend using the standard Long-CLIP strategy. 
In practice, we also observe a significant bias towards the first few tokens for SigLIP models (see Sec.~\ref{sec:32_biases_clip}), which further justifies freezing the first 20 embeddings.

\begin{table}[t]
\centering
\resizebox{0.47\textwidth}{!}
{\begin{tabular}{lcccc}
\toprule \multicolumn{1}{c}{\multirow{2}{*}{Method}} & COCO & Flickr & Urban1k & DOCCI \\
 & T2I & T2I & T2I & T2I \\
\midrule
\multicolumn{1}{l}{SigLIP} & & & & \\
 \quad Stretch all & 31.8 & 27.4 & 82.9 & 80.9 \\
 \quad \textbf{Freeze first 20} & 48.7 & 40.8 & 85.6 & 83.3  \\ 
\multicolumn{1}{l}{SigLIP2} & & & & \\
 \quad Stretch all & 39.5 & 28.3 & 87.0 & 82.9 \\
 \quad \textbf{Freeze first 20} & 51.9 & 43.6 & 87.2 & 83.0 \\ %
\bottomrule
\end{tabular}}
\caption{\textbf{Ablation on stretching all embeddings of SigLIP models for text-to-image (T2I) retrieval}. While performance on long retrieval (see Urban1k and DOCCI) does not change substantially, stretching all the embeddings leads to a significant reduction in short-text retrieval performance (see COCO and Flickr).
}
\label{tab:app_abl_stretch_siglip}
\end{table}

\subsubsection{Ablation on Number of Training Epochs}
We investigate the effects of training duration by training for 1, 3, 5, 7, and 10 epochs. Results are reported in \cref{tab:app_abl_num_epochs}. We observe that performance improves substantially across datasets when increasing from 1 to 3 epochs ($+2.9\%$ on Urban1k T2I), with more marginal but consistent gains from 3 to 5 and 5 to 7 epochs. Improvements for long retrieval from 7 to 10 epochs are minor ($0.2 \%$), and short retrieval is starting to be negatively affected. We also include the original Long-CLIP results for comparison. Notably, even with an equivalent training budget (1 epoch), our DeBias-CLIP method significantly outperforms the Long-CLIP baseline for all datasets, with $+1.8\%$ and $+1.9\%$ for short-text retrieval on COCO and Flickr T2I, and $+10.6\%$ and $+7.0\%$ for long-text retrieval on Urban1k and DOCCI T2I. 

\begin{table}[t]
\centering
\resizebox{0.47\textwidth}{!}
{\begin{tabular}{ccccc}
\toprule \multirow{2}{*}{Epochs} & COCO & Flickr & Urban1k & DOCCI \\
 & T2I & T2I & T2I & T2I \\
\midrule
1 (Long-CLIP) & 40.4 & 34.1 & 79.5 & 71.4 \\
3 (Long-CLIP) & 40.6 & 33.9 & 82.7 & 75.2 \\
\midrule
1  & 42.2 & 36.0 & 90.1 & 78.4 \\
\textbf{3}  & 43.0 & 36.6 & 93.0 & 80.0 \\
5  & 43.0 & 36.7 & 93.5 & 80.9 \\
7  & 43.2 & 36.8 & 94.1 & 81.3 \\
10 & 43.1 & 36.7 & 94.2 & 81.5 \\
\bottomrule
\end{tabular}}
\caption{\textbf{Ablation on the number of training epochs for text-to-image (T2I) retrieval}. We observe significant improvements when going from 1 to 3 epochs across all datasets. Extending training from 5 to 7 epochs yields smaller but still notable improvements on long retrieval (Urban1k and DOCCI). Results from 7 to 10 epochs are more marginal and mixed.
}
\label{tab:app_abl_num_epochs}
\end{table}

\subsubsection{Ablation on Sentence Sampling Method}
Our DeBias-CLIP method generates short captions by first generating a random number of sentences ${n_{\mrm{sampled}} = \mathcal{U}\left\{1\,,2\,,\dots,\,n_{\mrm{sents}}-1\right\}}$ from a uniform distribution, and then sampling that number without replacement from the original set of sentences $C^{\mrm{no\_sum}} = [s_{2},\,\dots,\,s_{n_{\mrm{sents}}}]$ (i.e., with the first sentence excluded), and finally concatenating the sentences together (\emph{Random} method). In this section, we consider four alternative sampling methods: 
\begin{itemize}
    \item \emph{Ordered}: We select a contiguous block of sentences starting from the second sentence up to a randomly selected end sentence, preserving the original narrative order.
    \item \emph{Independent}: We independently select each sentence in $C^{\mrm{no\_sum}}$ with a probability $p = 0.5$, creating a more uniform number of sentences.
    \item \emph{Keep 4}: We fix the sample size to $n_{\mrm{sampled}} = 4$ while sampling randomly without replacement.
    \item \emph{Shuffle}: We use the full set of sentences (no subsampling) but randomize their order.
\end{itemize}

Results are presented in \cref{tab:app_abl_sampling}. We find that performance differences between most sampling methods are marginal. Sampling contiguous sentences (\emph{Ordered}) improves short-text retrieval ($+1.0\%$ on Flickr) at the expense of long-text retrieval ($-1.3\%$ on Urban1k). This is similar to the SmartCLIP strategy, differing only in that SmartCLIP includes the initial summary sentence ($s_1$). Conversely, using all sentences (\emph{Shuffle}) boosts long retrieval ($+1.4\%$ on DOCCI) but causes significant degradation in short retrieval ($-1.3\%$ on Flickr). Ultimately, we find that our default \emph{Random} sampling offers the best balance between short and long retrieval without requiring additional parameter tuning. 


\begin{table}[t]
\centering
{\begin{tabular}{lcccc}
\toprule \multirow{2}{*}{Method} & COCO & Flickr & Urban1k & DOCCI \\
 & T2I & T2I & T2I & T2I \\
\midrule
\textbf{Random}  & 43.0 & 36.6 & 93.0 & 79.7 \\
Ordered & 43.2 & 37.6 & 91.7 & 80.2 \\
Independent  & 42.7 & 36.5 & 92.7 & 80.3 \\
Keep 4  & 42.7 & 36.4 & 92.8 & 80.0 \\
Shuffle & 41.7 & 35.3 & 92.8 & 81.1 \\
\bottomrule
\end{tabular}}
\caption{\textbf{Ablation on sentence sampling methods for text-to-image (T2I) retrieval}. We find that there is relatively little difference between different sampling methods. The \emph{Ordered} method improves short retrieval at the cost of long retrieval, while the \emph{Shuffle} method improves long retrieval on DOCCI at the cost of short retrieval. We find that our default \emph{Random} sampling offers the best tradeoff between short and long retrieval.
}
\label{tab:app_abl_sampling}
\end{table}

\subsubsection{Ablation on Padding Method}
In DeBias-CLIP, we add a random number of padding tokens before the tokenized text tokens from the caption, which depends on the post-caption padding length and leads to padding of varying lengths. Tab.~\ref{tab:abl_components} showed that our padding strategy (\emph{Random}) improves on using no padding (\emph{No pad}), and in Sec~\ref{sec:44_token_padding}, we hypothesize that one of the key benefits comes from shifting the short caption away from the initial token positions. As a result, early-position tokens are mostly trained with the full long caption,  and particularly its summary (first) sentence. In this ablation, we consider alternative fixed padding lengths of 20 and 40 tokens (\emph{Pad 20} and \emph{Pad 40}) and present results in \cref{tab:app_abl_padding}. Additionally, while our default method adds padding \textit{after} the \texttt{SOT} token (preserving its position), we also evaluate adding padding \textit{before} the \texttt{SOT} token (\emph{Pre-SOT}).

We find that all considered padding approaches improve short-text retrieval. Replacing random padding with fixed-length padding generally yields comparable performance, with only a marginal gain on DOCCI T2I ($+0.6\%$). Given this negligible difference, and to avoid introducing an arbitrary fixed hyperparameter, we retain the base random sampling approach. Finally, we find that adding padding after the \texttt{SOT} token generally outperforms padding before it (\emph{Pre-SOT}), justifying our default implementation. In \cref{sec:suppl_attn_weights} below, we discuss attention across tokens and highlight that the attention to the \texttt{SOT} token is much larger than for other tokens. Because the \texttt{SOT} token is present across all training datasets, is text-independent, and is always the first token, it plays a distinct role in the attention mechanism compared to regular text tokens. Moving it away from the first position could disrupt this, leading to worse performance. Recent work has shown that non-text positional tokens can serve as attention sinks in LLMs \cite{barbero2025llms}, reducing the magnitude of cross-attention between tokens when it isn't required. A similar effect could be present for CLIP text encoders.

\begin{table}[t]
\centering
{\begin{tabular}{lcccc}
\toprule \multirow{2}{*}{Method} & COCO & Flickr & Urban1k & DOCCI \\
 & T2I & T2I & T2I & T2I \\
\midrule
No pad & 41.9 & 35.8 & 92.5 & 80.8 \\
\textbf{Random}  & 43.0 & 36.6 & 93.0 & 79.7 \\
Pad 20  & 43.1 & 36.7 & 92.8 & 80.3 \\
Pad 40  & 43.0 & 36.7 & 92.8 & 80.3 \\
Pre-SOT  & 43.0 & 36.4 & 92.1 & 80.1 \\
\bottomrule
\end{tabular}}
\caption{\textbf{Ablation on token padding methods for text-to-image (T2I) retrieval}. All padding methods generally improve short-text performance. Replacing our default random padding (\emph{Random}) with fixed padding length of 20 (\emph{Pad 20}) or 40 (\emph{Pad 40}) yields comparable performance, with only marginal gains on DOCCI T2I ($+0.6\%$), while adding the padding before the \texttt{SOT} token leads to slightly worse performance.
}
\label{tab:app_abl_padding}
\end{table}

\subsubsection{Transferring to SmartCLIP}
In the main results, we show that the text sampling used in DeBias-CLIP substantially outperforms the Long-CLIP baseline. Here, we apply similar sampling methods to SmartCLIP \cite{xie2025smartclip}, which was consistently one of the best-performing methods for both long and short retrieval (see Sec.~\ref{sec:53_results}). SmartCLIP trains with a single caption obtained by truncating the original long caption after a random number of sentences. Thus, it always contains the summary sentence. As alternatives, we consider two sampling methods: one identical to SmartCLIP except that we always remove the summary sentence by starting from the second sentence (\emph{Start at 2}), and one where we keep the whole caption except the first sentence (\emph{Only drop 1}). We train with SmartCLIP's released code, use a ViT-B/16 backbone, and run for 3 epochs. Results are presented in \cref{tab:app_abl_smartclip}. We find that removing the first sentence consistently improves long-text retrieval performance (\emph{Original} vs \emph{Start at 2}, $+2.7\%$ on Urban1k T2I), with mixed results on short-text retrieval,  with$+0.7\%$ on COCO T2I but $-0.5\%$ for I2T (not in the table), and better performance on Flickr. We can further improve long retrieval by not subsampling the caption sentences (\emph{Only drop 1}), at the cost of some short-text performance. However, the final short retrieval results remain similar to the original SmartCLIP while being much better ($+5.1\%$ on Urban1k T2I) for long-text retrieval. Overall, our results on SmartCLIP highlight the importance of not training with the summary sentence. 

\begin{table}[t]
\centering
{\begin{tabular}{lcccc}
\toprule \multirow{2}{*}{Method} & COCO & Flickr & Urban1k & DOCCI \\
 & T2I & T2I & T2I & T2I \\
\midrule
Original & 42.4 & 36.3 & 87.4 & 78.0 \\
Start at 2  & 43.1 & 37.4 & 90.1 & 79.7 \\
Only drop 1 & 42.2 & 36.3 & 92.5 & 81.8 \\
\bottomrule
\end{tabular}}
\caption{\textbf{Ablation on caption sampling with the SmartCLIP training on text-to-image (T2I) retrieval.} We obtain consistently better performance compared to the original SmartCLIP by not sampling the first (summary) sentence. There are clear tradeoffs between short-text (COCO and Flickr) and long-text (Urban1k and DOCCI) retrieval when subsampling the sentences instead of keeping the full original caption.
}
\label{tab:app_abl_smartclip}
\end{table}

\subsection{Additional Discussion on CLIP Biases}
\label{sec:suppl_additional_clip_biases}

\begin{figure*}[ht]
    \centering
    \subcaptionbox{DOCCI T2I \label{fig:abl_swap_docci_t2i.pdf}}[0.223\textwidth]{\includegraphics[width=0.223\textwidth]{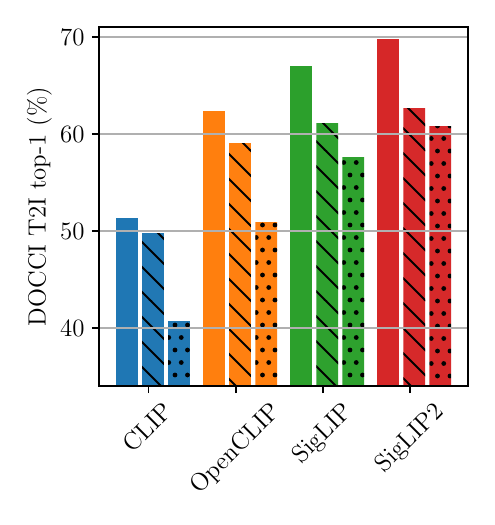}}
    \subcaptionbox{DOCCI I2T \label{fig:abl_swap_docci_i2t.pdf}}[0.223\textwidth]{\includegraphics[width=0.223\textwidth]{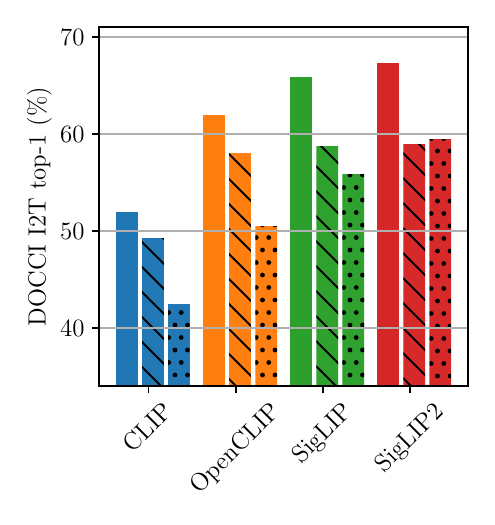}}
    \subcaptionbox{Urban1k T2I \label{fig:abl_swap_urban1k_t2i.pdf}}[0.223\textwidth]{\includegraphics[width=0.223\textwidth]{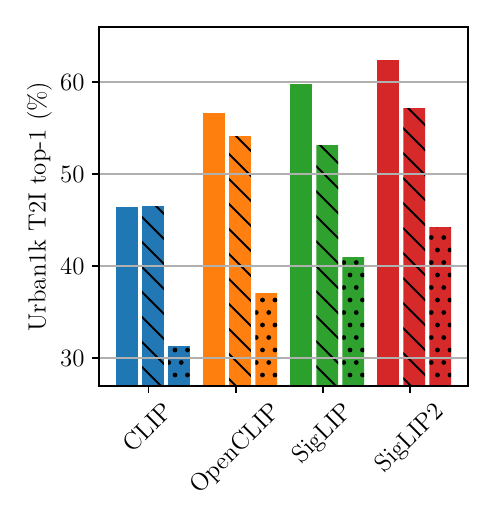}}
    \subcaptionbox{Urban1k I2T \label{fig:abl_swap_urban1k_i2t.pdf}}[0.223\textwidth]{\includegraphics[width=0.223\textwidth]{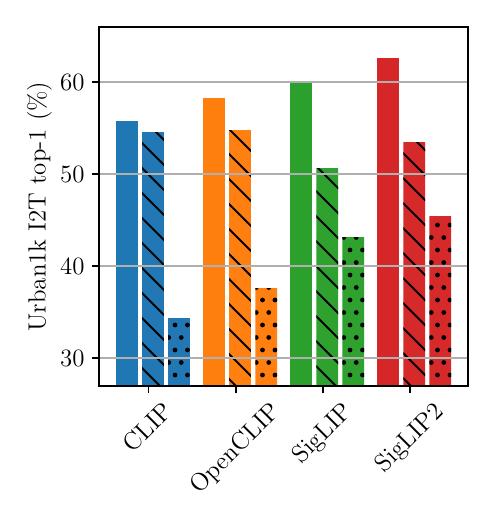}}
    \begin{subfigure}[b]{0.078\textwidth}{\includegraphics[width=\textwidth]{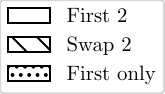}} \vspace{45pt}
    \end{subfigure}
\caption{\textbf{Effects of permutation of the first two sentences on the top-1 text-to-image (T2I) and image-to-text (I2T) retrieval performance for CLIP models.} We present results on DOCCI in (a) and (b), and on Urban1k in (c) and (d). We analyze three setups: the first two sentences in the correct order (\emph{First 2}), the same two sentences swapped (\emph{Swap 2}), and the first sentence only (\emph{First only}). Sentence swapping generally leads to worse retrieval performance and tends to be more severe for the SigLIP and SigLIP2 models. Dropping the second sentence leads to significantly larger degradation on Urban1k compared to DOCCI.}
\label{fig:abl_swap_docci_urban1k}
\end{figure*}

\begin{figure*}[ht]
    \centering
    \subcaptionbox{DOCCI T2I \label{fig:abl_padding_docci_t2i.pdf}}[0.223\textwidth]{\includegraphics[width=0.223\textwidth]{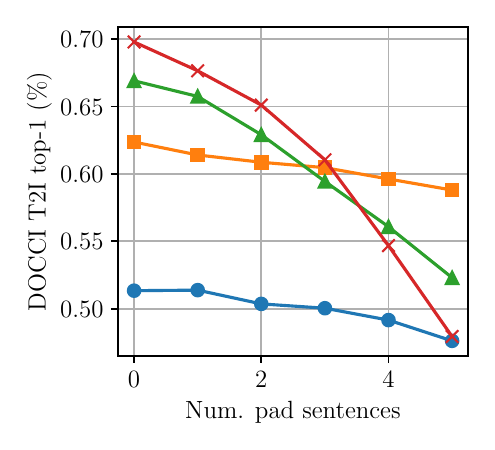}}
    \subcaptionbox{DOCCI I2T \label{fig:abl_padding_docci_i2t.pdf}}[0.223\textwidth]{\includegraphics[width=0.223\textwidth]{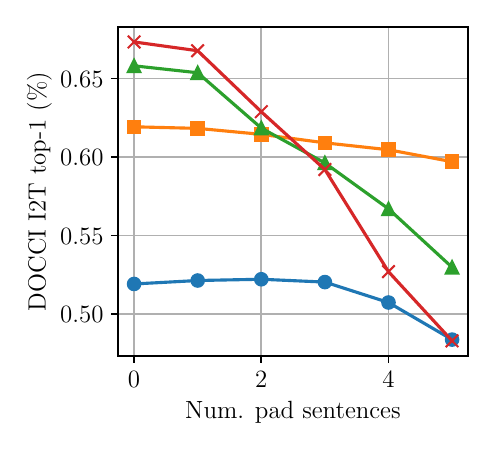}}
    \subcaptionbox{Urban1k T2I \label{fig:abl_padding_urban1k_t2i.pdf}}[0.223\textwidth]{\includegraphics[width=0.223\textwidth]{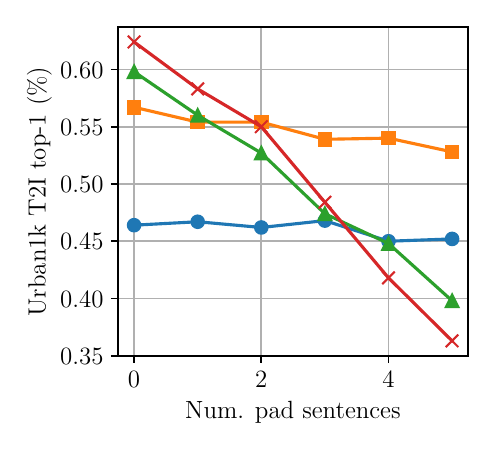}}
    \subcaptionbox{Urban1k I2T \label{fig:abl_padding_urban1k_i2t.pdf}}[0.223\textwidth]{\includegraphics[width=0.223\textwidth]{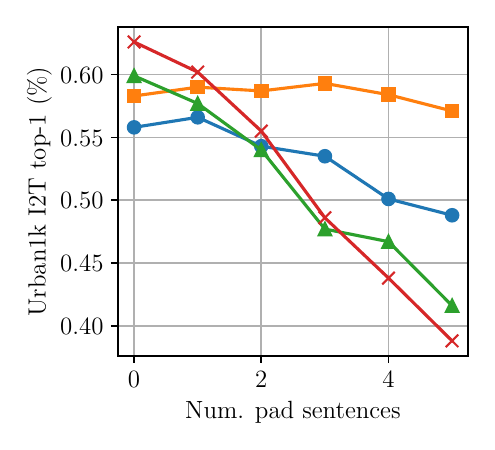}}
    \begin{subfigure}[b]{0.078\textwidth}{\includegraphics[width=\textwidth]{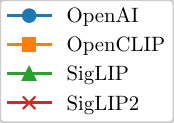} \vspace{30pt}}
    \end{subfigure}
\caption{\textbf{Effects of padding sentences on the top-1 text-to-image (T2I) and image-to-text (I2T) retrieval performance for CLIP models.} We present results on DOCCI in (a) and (b), and on Urban1k in (c) and (d). One to five padding sentences \texttt{`This is a photo.'} are added before the truncated original caption (we keep the first two sentences only). Increased padding consistently leads to worse retrieval performance, but this effect is much more severe for the SigLIP and SigLIP2 models.}
\label{fig:abl_padding_docci_urban1k}
\end{figure*}

We extend the empirical analysis from Sec.~\ref{sec:3_empirical_analysis} with additional results on DOCCI and Urban1k. We continue our analysis of retrieval with the first two sentences of long-caption datasets (see Sec.~\ref{sec:32_biases_clip}), comparing three configurations: using both sentences in order (\emph{First 2}), both sentences with the order swapped (2 before 1, \emph{Swap 2}), and using only the first sentence (\emph{First only}). For DOCCI, the average token count for the first two sentences is 46.6, 43.8, and 40.4 for CLIP, SigLIP, and SigLIP2 tokenizers, respectively. For Urban1k, these averages are 49.5, 46.5, and 42.1, respectively. Results are presented in \cref{fig:abl_swap_docci_urban1k}. When comparing models, we broadly observe the same trends, with significant performance drops in the \emph{Swap 2} case, particularly for SigLIP and SigLIP2. We observe slightly different trends between retrieval on DOCCI and Urban1k. OpenAI CLIP is less sensitive to sentence permutation on Urban1k ($+0.1\%$ and $-1.3\%$ for Urban1k T2I and I2T, respectively), but this could be due to lower performance in general. Furthermore, performance for the \emph{First only} setting on Urban1k is substantially worse than the \emph{First 2} baseline ($-9.5\%$ on DOCCI I2T and $-21.5\%$ on Urban1k I2T for OpenAI CLIP), indicating that additional text is critical for retrieval on this dataset. This could be due to Urban1k images being mostly limited to ground-level urban scenes, compared to the higher diversity of DOCCI. Consequently, many Urban1k captions likely share similar first (summary) sentences, making the second sentence essential for disambiguation. 

\cref{fig:abl_padding_docci_urban1k} presents additional results for the padding effects on DOCCI and Urban1k. We see similar trends across models and datasets, with CLIP models (OpenAI and OpenCLIP) exhibiting significantly less sensitivity to padding compared to SigLIP models. While SigLIP models perform better than CLIP variants in the no-padding setup, this advantage does not hold for longer padding sequences. This aligns with our results for sentence transposition and implies that SigLIP models have a very narrow effective context window.

\subsection{Additional Discussion on Long-CLIP Biases}
\label{sec:suppl_additional_longclip_biases}

\subsubsection{Long-Text Retrieval with Sentence Permutation and Removal}
\begin{figure*}[ht]
    \centering
    \subcaptionbox{DOCCI T2I ViT-L \label{fig:abl_long_docci_t2i.pdf}}[0.223\textwidth]{\includegraphics[width=0.223\textwidth]{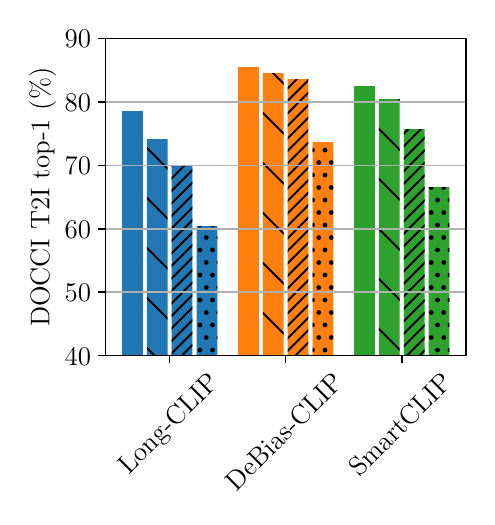}}
    \subcaptionbox{DOCCI I2T ViT-L \label{fig:abl_long_docci_i2t.pdf}}[0.223\textwidth]{\includegraphics[width=0.223\textwidth]{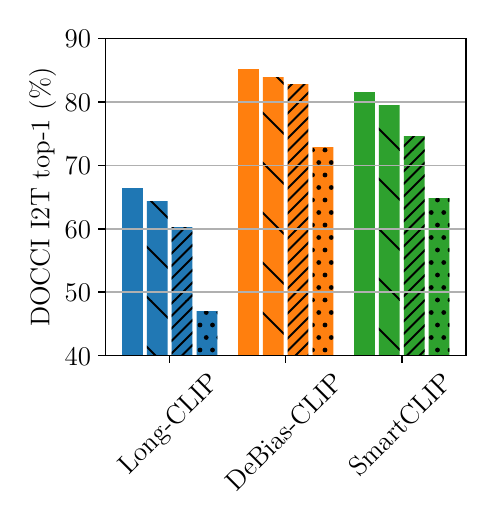}}
    \subcaptionbox{Urban1k T2I ViT-L \label{fig:abl_long_urban1k_t2i.pdf}}[0.223\textwidth]{\includegraphics[width=0.223\textwidth]{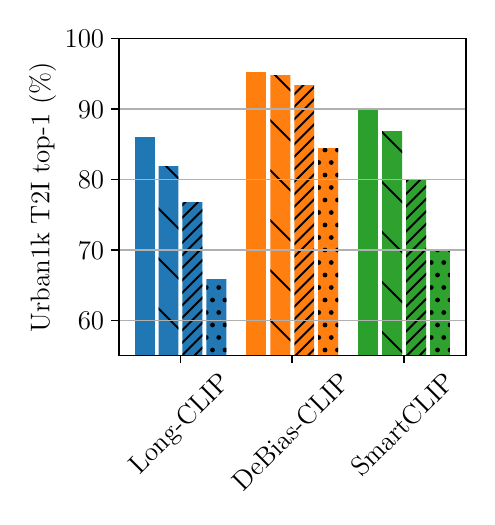}}
    \subcaptionbox{Urban1k I2T ViT-L \label{fig:abl_long_urban1k_i2t.pdf}}[0.223\textwidth]{\includegraphics[width=0.223\textwidth]{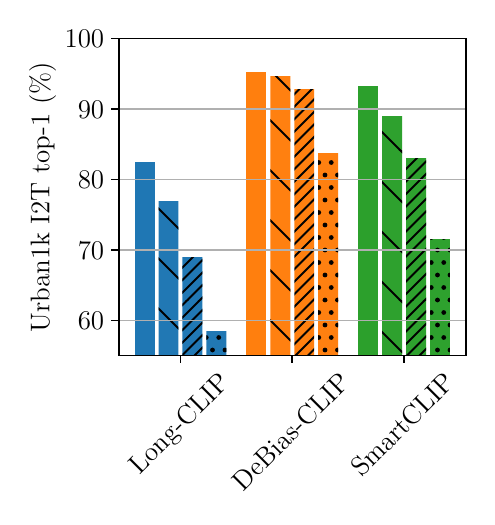}}
    \begin{subfigure}[b]{0.078\textwidth}{\includegraphics[width=\textwidth]{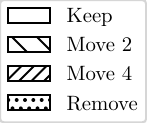}} \vspace{45pt}
    \end{subfigure}
\caption{\textbf{Comparison of long-text top-1 retrieval with sentence permutations for Long-CLIP \cite{zhang2024long}, SmartCLIP \cite{xie2025smartclip} and our proposed method (DeBias-CLIP).} We consider four cases: the original full caption (\emph{Keep}), transposing the first and second sentences (\emph{Move 2}), transposing the first and fourth sentences (\emph{Move 4}), and removing the first sentence (\emph{Remove}). We present results on DOCCI in (a) and (b), and on Urban1k in (c) and (d). Sentence permutation generally leads to worse retrieval performance, but our method demonstrates greater robustness compared to either Long-CLIP or SmartCLIP.}
\label{fig:abl_long_docci_urban1k}
\end{figure*}

In sections~\ref{sec:33_biases_longclip} and \ref{sec:53_results}, we showed that our DeBias-CLIP method achieves better long-text retrieval performance and is more robust than Long-CLIP to sentence transposition or to erasing the first sentence. Here, we extend the comparison to SmartCLIP, which is a consistently strong model for both long- and short-text retrieval. We repeat the text permutation analysis from Sec.~\ref{sec:533_gen_to_other_CLIP} but now consider ViT-L models and additionally evaluate on both DOCCI and Urban1k to show that these results generalize broadly. We consider four text augmentation cases: the original full caption (\emph{Keep}), transposing the first and second sentences (\emph{Move 2}), transposing the first and fourth sentences (\emph{Move 4}), and removing the first sentence (\emph{Remove}). If there are fewer than four sentences, we transpose the last sentence with the first. We present results in \cref{fig:abl_long_docci_urban1k}. First, we note that DeBias-CLIP is consistently better in the nominal case (\emph{Keep}) and substantially more robust to sentence permutations. On DOCCI T2I, the \emph{Move 4} case leads to a reduction of $-8.7\%$ for Long-CLIP and $-6.9\%$ for SmartCLIP, but only $-2.0\%$ for DeBias-CLIP. Even when removing the first sentence, DeBias-CLIP loses only $-12.1\%$, while Long-CLIP and SmartCLIP lose $-18.3\%$ and $-15.9\%$, respectively, on DOCCI T2I. On Urban1k T2I, DeBias-CLIP loses only $-10.8\%$ compared to $-20.2\%$ for Long-CLIP and $-20.1\%$ for SmartCLIP, showing that our model is significantly better at using details in the later sentences of the caption to match images without relying on the summary sentence information.

\subsubsection{Encoder Design to Resolve the Early Token Bias}
Our method resolves the early token bias issue from a data perspective by avoiding training with the summary captions. In this section, we investigate two possible encoder design solutions to reduce the bias: using relative positional embedding instead of the absolute positional embedding, and using token average pooling as the output token instead of aggregating information to a single token.

\begin{figure}[ht]
  \centering
    \includegraphics[width=0.40\textwidth]    {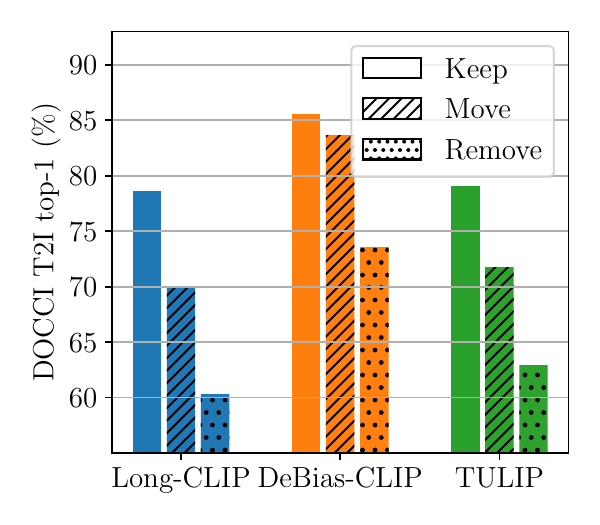}
        \caption{\textbf{Effects of permutations of the first two sentences on top-1 text-to-image retrieval for TULIP ViT-L.} While TULIP is consistently better than Long-CLIP, it still shows significant performance degradation when sentences are permuted (\emph{Move}) or the first sentence is removed (\emph{Remove}).}
    \label{fig:abl_vs_tulip_permutations}
\end{figure}

TULIP \cite{najdenkoska2024tulip} proposes to learn relative positional embedding, particularly rotary positional embeddings (RoPE) \cite{su2024roformer}, instead of linearly interpolating the original CLIP absolute positional embeddings like Long-CLIP. RoPE is commonly used in language models and has been shown to have better interpolation and extrapolation properties. TULIP shows significant improvements over Long-CLIP, but it is unclear whether this is because the method resolves the early token bias or because the model is generally better. Thus, we evaluate TULIP ViT-L in the sentence permutation setting considered in \cref{sec:534_sentence_permutation} (\emph{Move} permute first and fourth sentences, \emph{Remove} remove first sentence from caption) and show results in \cref{fig:abl_vs_tulip_permutations}. While TULIP is consistently better than Long-CLIP, performance is significantly worse when the first sentence is moved ($-7.3\%$) or removed ($-16.1\%)$ when compared to our DeBias-CLIP ($-2.0\%$ and $-12.1\%$ respectively). This shows that the bias problem persists even when using relative positional embeddings.

\begin{figure}[ht]
  \centering
    \includegraphics[width=0.40\textwidth]    {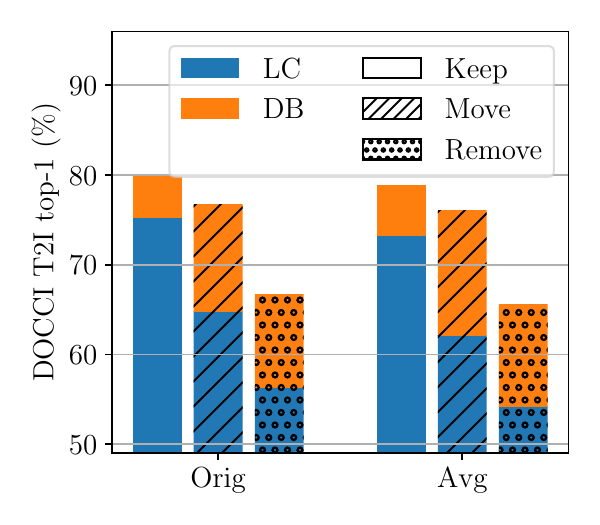}
        \caption{\textbf{Effects of permutations of the first two sentences on top-1 text-to-image retrieval with token average pooling for Long-CLIP (LC) and DeBias-CLIP (DB).} Token average pooling generally does slightly worse, and does not reduce early token bias.}
    \label{fig:abl_avgpool_permutations}
\end{figure}

Next, we consider using token average pooling to generate the final text output token, compared to aggregating information in a single token (the \texttt{EOT} token for CLIP). This would force the final text output to be explicitly dependent on all the encoded tokens, potentially making it more sensitive to information later in the caption. We train both Long-CLIP and DeBias-CLIP ViT-B for 3 epochs with token averaging (\textit{Avg}) and again consider the sentence permutation of \cref{sec:534_sentence_permutation}, comparing them to the original CLIP single token aggregation (\textit{Orig}) in \cref{fig:abl_avgpool_permutations}. We see that average pooling generally reduces performance by a small amount (about $-2\%$ for Long-CLIP, $-1\%$ for DeBias-CLIP) and does not resolve the issue of early token bias. For the original Long-CLIP, we see a drop of $-10.4\%$ when permuting the first and fourth sentences (\textit{Move}), and a drop of $-11.0\%$ with average pooling.

\subsubsection{Additional Details and Results on the Attention Weights Distribution}
\label{sec:suppl_attn_weights}
\begin{figure}[ht]
  \centering
    \includegraphics[width=0.46\textwidth]    {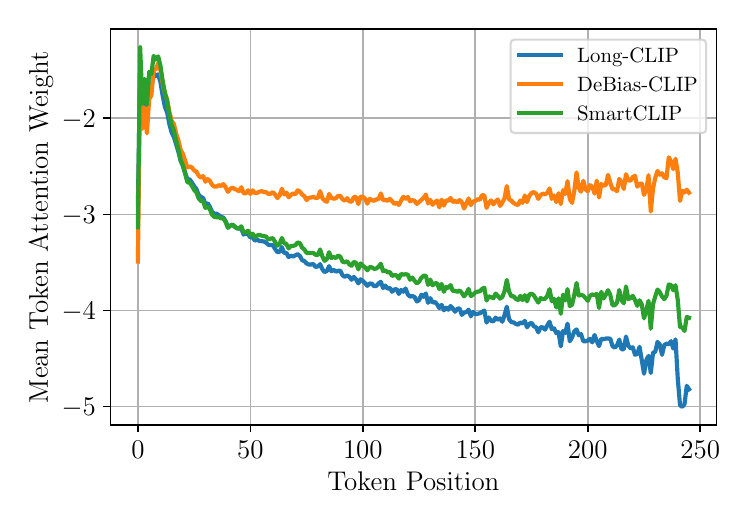}
        \caption{\textbf{Mean \emph{pre-softmax} self-attention weight from the output token to the text tokens as a function of the token position on DOCCI.} For both Long-CLIP and SmartCLIP, the attention magnitude consistently decays beyond 50 tokens, while for DeBias-CLIP it remains essentially flat (ignoring high-variance values at the very end).}
    \label{fig:abl_mean_token_attn_smart}
\end{figure}

\begin{figure}[ht]
  \centering
    \includegraphics[width=0.46\textwidth]    {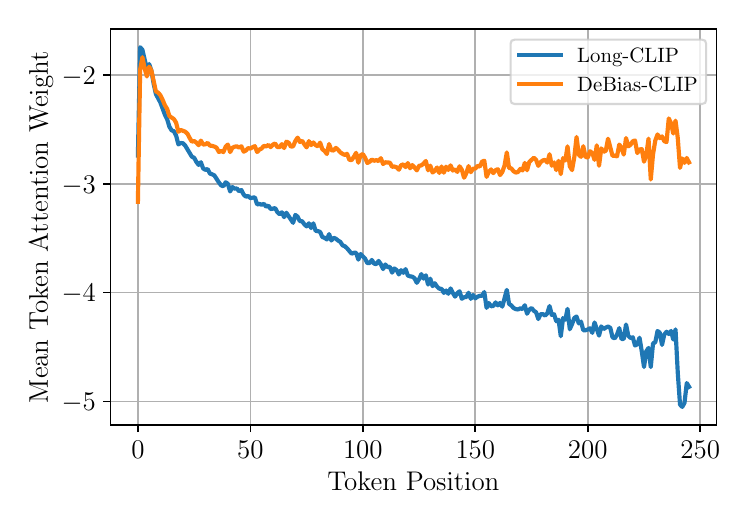}
        \caption{\textbf{Mean \emph{pre-softmax} self-attention weight from the output token to the text tokens as a function of the token position on DOCCI, with the first and fourth sentences transposed.} We see similar trends as in the untransposed case: there is a significant bias toward early tokens even if the summary sentence is moved to a later position. The summary sentence visibly receives more attention even after being moved (around tokens 50-100).}
    \label{fig:abl_mean_token_attn_swap}
\end{figure}

\begin{figure}[ht]
  \centering
    \includegraphics[width=0.46\textwidth]    {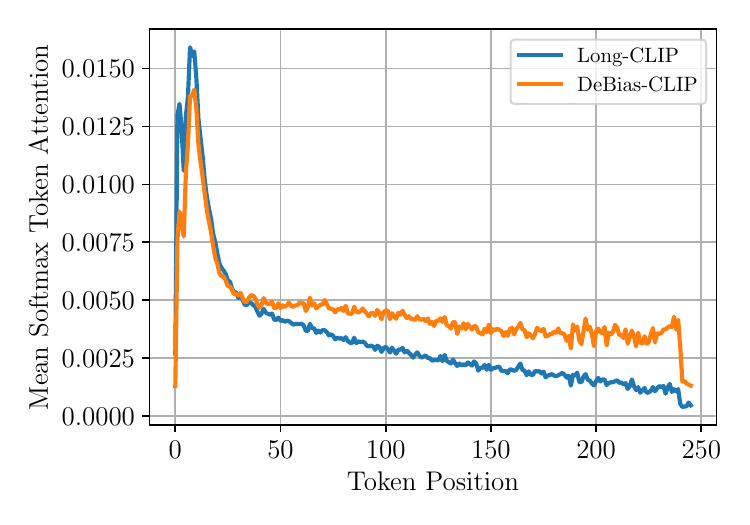}
        \caption{\textbf{Mean \emph{post-softmax} self-attention weight from the output token to the text tokens as a function of the token position on DOCCI.} DeBias-CLIP reduces the bias on early tokens and has a relatively even attention spread on later tokens, while we observe a decay for Long-CLIP.}
    \label{fig:abl_sm_token_attn}
\end{figure}

In Fig.~\ref{fig:placeholder}~(b), we present a plot of the average attention weights over tokens for both Long-CLIP and our DeBias-CLIP method. We provide additional details here. Our goal is to show that Long-CLIP models are biased towards early tokens in the caption. To this end, we measure the token self-attention values in the attention matrix \emph{before} the softmax, and consider specifically the attention from the output token (analogous to the \texttt{CLS} token for the image encoder and it is the \texttt{EOT} token for the CLIP text encoder) to all other positional tokens in the last transformer layer, which captures how much information from the other positions is used to update the output token. We use the pre-softmax values as these directly capture the token-to-token affinity without the effects of the softmax normalization.

For a single caption, the query $\mbf{Q} \in \mathbb{R}^{D}$ and key $\mbf{K} \in \mathbb{R}^{D}$ matrices are used in self-attention (before softmax) as
\begin{align}
    \mbf{M} &= \left(\dfrac{\mbf{Q}\mbf{K}^{\trans}}{\sqrt{D}}\right),
\end{align}
where $\mbf{M} \in \mathbb{R}^{D \times D}$ is the self-attention weight matrix across all token positions and $D$ is the hidden dimension. We are interested in the row of matrix $\mbf{M}$ corresponding to the output token (i.e., \texttt{EOT} position) $\mbf{m}_{\mathtt{EOT}}$ given by
\begin{align}
    \mbf{m}_{\mathtt{EOT}} &= \dfrac{\mbf{q}_{\mrm{EOT}} \mbf{K}^{\trans}}{\sqrt{D}},
\end{align}
which is the vector of attention weights from the output token to all other tokens. Because CLIP models use a causal mask in the text encoder, padding tokens that appear after the \texttt{EOT} token have an attention weight of $-\infty$ (this becomes a weight of 0 after the softmax), so $\mbf{m}_{\mathtt{EOT}}$ has the form ${\mbf{m}_{\mathtt{EOT}} = [v_{\mathtt{SOT}},v_1,\dots,v_{k},v_{\mathtt{EOT}},-\infty,\dots,-\infty]}$. 

In Fig.~\ref{fig:placeholder}~(b), we plot this value aggregated over the DOCCI dataset. For each token position,  we sum the corresponding entries of $\mbf{m}_{\mathtt{EOT}}$ over all captions and divide by the number of times that position corresponds to a text token (rather than post-text padding), giving a normalizing vector $\mbf{n}$: 
\begin{align}
    \mbf{n} &= \left[n_1,\,n_2,\,\dots,\,n_{248}\right] \nonumber \\
    &= \sum_{i=1}^{N} \mbf{m}_{\mathtt{EOT},i} > -\infty,
\end{align}
where $N$ is the number of samples in the dataset, and where the inequality is applied independently for each element of $\mbf{m}_{\mathtt{EOT},i}$. This yields a vector of binary elements for each caption in the dataset. The normalization excludes the padding tokens, which are more frequent for later embedding positions, and captures the average value of attention weights when they correspond to text tokens. In Fig.~\ref{fig:placeholder}~(b) and in the following figures, we exclude the weight with respect to the \texttt{SOT} token $v_{\mathtt{SOT},i}$, which is generally much larger (on the order of 2.5), and corresponds to a softmax probability between 0.35 and 0.40.  

Similar to how our method improves Long-CLIP by assigning more consistent attention weights to later text tokens, \cref{fig:abl_mean_token_attn_smart} repeats the analysis for SmartCLIP, showing it has a less steep drop in attention over token positions compared to Long-CLIP, which is consistent with its improved long-text retrieval performance. However, there is still substantially less attention on later tokens in SmartCLIP. Finally, when evaluated on DOCCI, where the first $\sim$25 tokens correspond to the summary sentence, we see that the earliest text positions consistently have larger attention weights on average, even for DeBias-CLIP. This could be related to a positional bias toward earlier tokens and/or a bias towards the summary sentence. 

To disentangle these effects, we replicate the experiment under the \emph{Move 4} setting, where we transpose the first and fourth sentences. Results in \cref{fig:abl_mean_token_attn_swap} show that (i) a positional bias remains, with significantly larger attention values for early tokens, and (ii) the summary sentence continues to receive more attention (as it likely contains more information) even when moved to a later position. In fact, we see an increase of about 0.2 in the attention values from token positions 50 to 100 in the \emph{Move 4} setting. Additionally, we present the post-softmax attention weights in \cref{fig:abl_sm_token_attn}. After softmax normalization, the larger weights on later tokens in DeBias-CLIP lead to a redistribution of the attention from the first few tokens to the rest of the caption.

\begin{figure*}[t]
  \centering
    \includegraphics[width=0.95\textwidth]
    {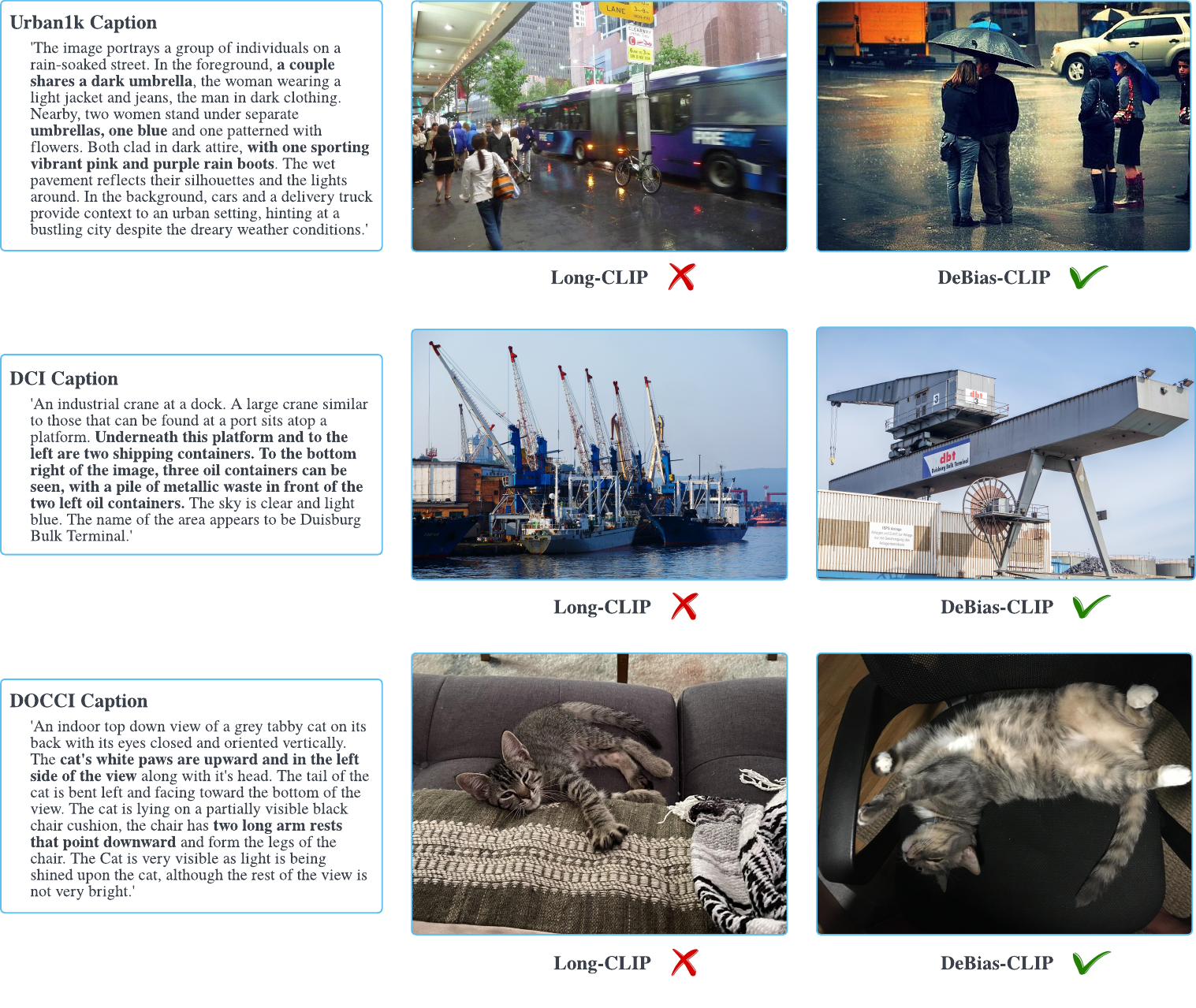}
    	\caption{\textbf{Comparison between Long-CLIP and DeBias-CLIP on long text-to-image retrieval.} We observe that DeBias-CLIP more consistently recovers images that are correctly matched to the text. Text in bold marks the caption details that are required to retrieve the right image.}
    	\label{fig:comparison}
\end{figure*}

\subsection{Visualizations and Results for Image Generation}
\label{sec:suppl_visualizations}
We claim that DeBias-CLIP is more sensitive to caption details than Long-CLIP, particularly in later sentences. \cref{fig:comparison} presents examples on Urban1k, DCI, and DOCCI, where Long-CLIP fails to retrieve the correct image given the long caption, but where DeBias-CLIP succeeds. In general, these are cases where the first summary caption applies equally to either image, and where specific details in the caption (e.g., presence of umbrellas, oil containers, the color of the cat's paws) are required to identify the matching image.

\begin{figure*}[t]
  \centering
    \includegraphics[width=0.95\textwidth]
    {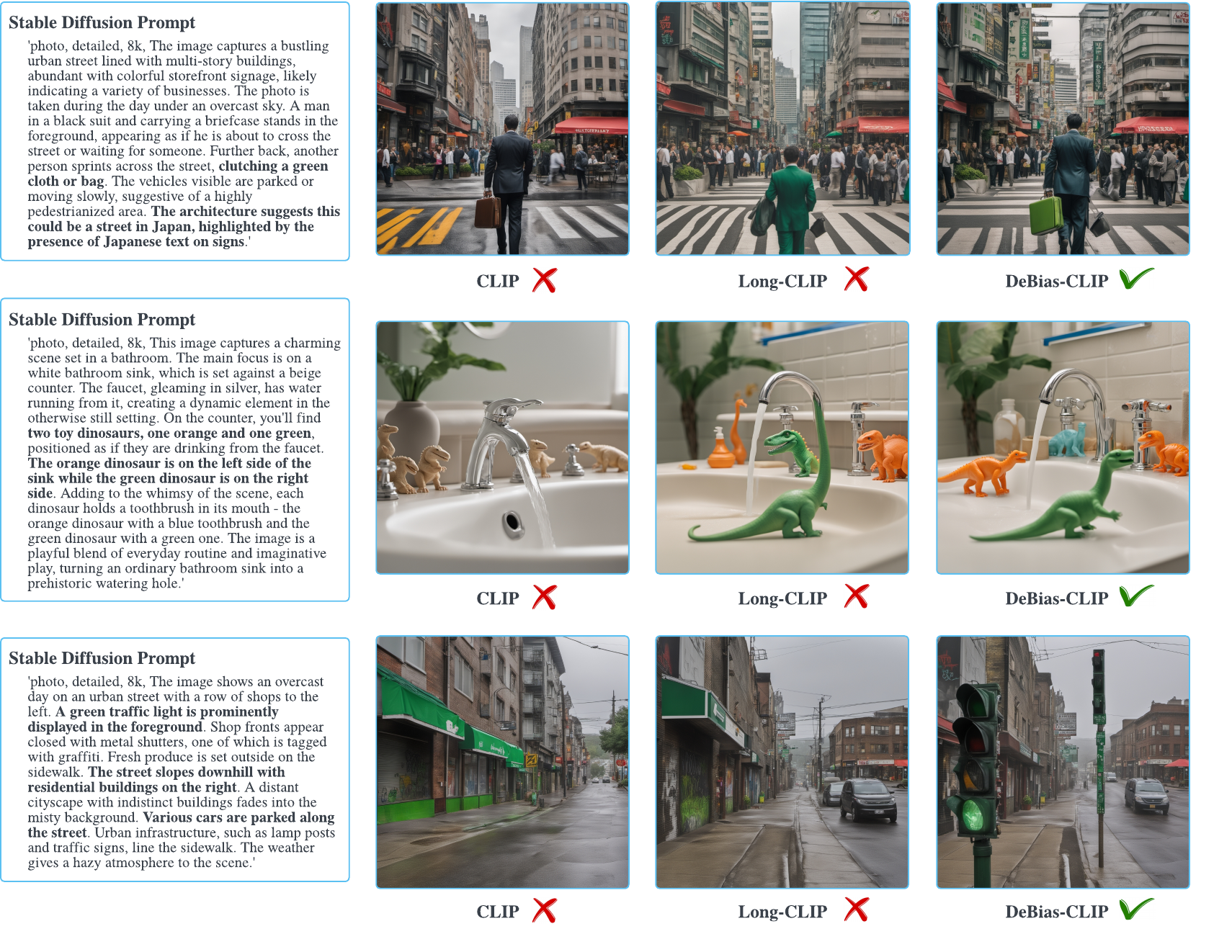}
    	\caption{\textbf{Stable Diffusion text-to-image generation results for CLIP, Long-CLIP, and DeBias-CLIP.} DeBias-CLIP can leverage more details that appear later in the caption and suffers less from inaccurate localization of details (e.g., the green color of the bag being transferred to the suit in the Long-CLIP first row image). Text in bold highlights the caption details that are correctly represented in our DeBias-CLIP method.}
    	\label{fig:stable_diff}
\end{figure*}

We also investigate how our DeBias-CLIP models perform on downstream tasks. In particular, we replace the CLIP text encoder from SDXL \cite{podellsdxl} with DeBias-CLIP. \cref{fig:stable_diff} compares text-to-image generation with the original CLIP, Long-CLIP, and DeBias-CLIP, with prompts from ShareGPT4V and Urban1k captions. The generation results show that CLIP struggles with long captions and often fails to capture key details, such as the Japanese text in the first example, or the color of the dinosaurs in the second. When comparing DeBias-CLIP to Long-CLIP, we find that DeBias-CLIP preserves more fine-grained details (e.g., the green traffic light in the third example). However, generation with long captions also highlights a broader limitation of all CLIP models: they suffer from poor relational understanding, with limited ability to encode relative position (left/right), accurate object counts, and consistent object-attribute matches. For example, the blue dinosaur generated by DeBias-CLIP in row 2 is an incorrect amalgamation of an orange dinosaur holding a blue toothbrush. Quantitatively, we compare the performance of DeBias-CLIP with Long-CLIP by measuring the similarity between the original images and images generated with the 1000 paired long caption from the ShareGPT4v validation split. We evaluate Fréchet inception distance (FID) \cite{heusel2017gans} and the DINO score \cite{caron2021emerging}. \cref{tab:sdxl_quant} shows that DeBias-CLIP improves both metric, showing we generate images from the long captions that are semantically closer to the paired images.

\begin{table}[h!]
\centering
{\begin{tabular}{ccc}
\toprule
Model & FID$\downarrow$ & DINO Score$\uparrow$ \\
\midrule
Long-CLIP & 0.335 & 0.515 \\
DeBias-CLIP (ours) & 0.317 & 0.528 \\
\bottomrule
\end{tabular}}
\caption{\textbf{Quantitative image generation comparison}. We replace the CLIP text encoder in SDXL with either Long-CLIP or DeBias-CLIP  generate images from the 1000 image-caption pairs of the ShareGPT4V validation set. DeBias-CLIP generates images that are semantically more similar to the original images.
}
\label{tab:sdxl_quant}
\end{table}

\end{document}